\documentclass{article}

\usepackage[preprint]{corl_2026} 
\usepackage{graphicx}
\usepackage{tabularx} 
\usepackage{booktabs}
\usepackage{multirow}
\usepackage{amsmath}  
\usepackage{amssymb}
\usepackage{wrapfig}
\usepackage[table]{xcolor}
\usepackage{hyperref}
\usepackage{siunitx}
\title{Embody4D: A Generalist Data Engine for Embodied 4D World Modeling}

\author{
\begin{tabular}{c}
{\small\textbf{
Peiyan Tu\textsuperscript{1,2,*},
Hanxin Zhu\textsuperscript{3,*},
Jingwen Sun\textsuperscript{3,2},
Shaojie Ren\textsuperscript{4,2},
Cong Wang\textsuperscript{4,2},}}\\
{\small\textbf{
Yuyan Xu\textsuperscript{5,2},
Jiayi Luo\textsuperscript{6,2},
Xiaoqian Cheng\textsuperscript{3,2},
Zhibo Chen\textsuperscript{3,2,\textdagger}}}\\[0.4em]
{\footnotesize
\textsuperscript{1}Zhejiang University,
\textsuperscript{2}Beijing Zhongguancun Academy,
\textsuperscript{3}University of Science and Technology of China,}\\
{\footnotesize
\textsuperscript{4}Institute of Automation, Chinese Academy of Sciences,
\textsuperscript{5}Shanghai Jiao Tong University,}\\
{\footnotesize
\textsuperscript{6}Beihang University}\\[0.2em]
{\footnotesize
\textsuperscript{*}Equal contribution.
\textsuperscript{\textdagger}Corresponding author.
}
\end{tabular}
}

\begin{document}
\maketitle

\vspace{-0.5cm}
\begin{abstract}

Embodied agents require robust and comprehensive 3D spatiotemporal representations to support spatial reasoning, manipulation understanding, and downstream decision making. However, existing robot data are typically captured from fixed or sparse viewpoints, providing only partial and view-dependent observations, which limits multi-view perception and generalization across viewpoints. Given the difficulty of collecting additional viewpoints in real-world settings, we propose Embody4D, a dedicated video-to-video world model for embodied scenarios to bridge this observation gap by transforming a monocular robot video into novel-view videos from flexible target camera viewpoints. First, to tackle training data scarcity, we introduce a 3D-aware compositional synthesis pipeline to curate a heterogeneous dataset compositing cross-embodiment robotic arms with diverse backgrounds, promoting broad generalization. Second, to enforce geometric stability, we devise a latent confidence-aware expert modulation strategy, which estimates the reliability of warped latent priors and adaptively routes regions to copy, repair, or inpaint experts for spatiotemporally consistent 4D generation. Finally, to enhance the fidelity of the manipulation, we incorporate an interaction-aware attention mechanism that explicitly attends to the robotic interaction regions. Extensive experiments show that Embody4D achieves state-of-the-art performance on visual evaluation benchmarks, while both simulated and real-world robotic experiments further demonstrate its effectiveness as a robust data engine for synthesizing high-fidelity, view-consistent videos that empower downstream robotic planning and learning. The anonymous project page: \url{https://peiyantu.github.io/Embody4D/page.html}
\end{abstract}



\section{Introduction}
\label{sec:intro}
    Embodied agents require robust and comprehensive 3D spatiotemporal representations to support spatial reasoning, manipulation understanding, and downstream decision making. However, while the physical world is inherently three-dimensional, many existing models for downstream policy learning remain confined to 2D pixel observations \cite{openvla,0.5} or rely on limited, fixed multi-view captures \cite{dreamzero}. As a result, they struggle to generalize across unseen viewpoints or construct complete 3D spatiotemporal scene representations. This limitation results in impoverished spatial representations, depriving embodied agents of the comprehensive multi-view information required for robust manipulation and spatial reasoning \cite{pointworld}.

    A key bottleneck behind this limitation is the scarcity of dense multi-view video data with flexible viewpoint coverage in embodied scenarios \cite{4ddata}. Although simulators offer scalable access to controllable multi-view observations, the resulting data often remain misaligned with real-world robot interactions, particularly in appearance, dynamics, and contact-rich behaviors \cite{robotwin}. In contrast, acquiring dense multi-view videos in the real world requires carefully calibrated and synchronized camera systems, making large-scale collection prohibitively expensive and difficult to deploy \cite{agibot}.

    This motivates a generalist data engine for transforming widely available monocular or sparsely captured robot videos into dense multi-view videos under flexible camera viewpoints. Such videos faithfully capture the same robot-object interaction across viewpoints and time, forming a temporally coherent multi-view representation of the dynamic 3D scene. We refer to this representation as 4D embodied data, which provides embodied agents with dense spatiotemporal supervision for cross-view generalization, scene understanding, and manipulation reasoning \cite{4dgeneration:Asurvey}.

\begin{figure}[t]
        \raggedright
        \includegraphics[width=0.95\linewidth]{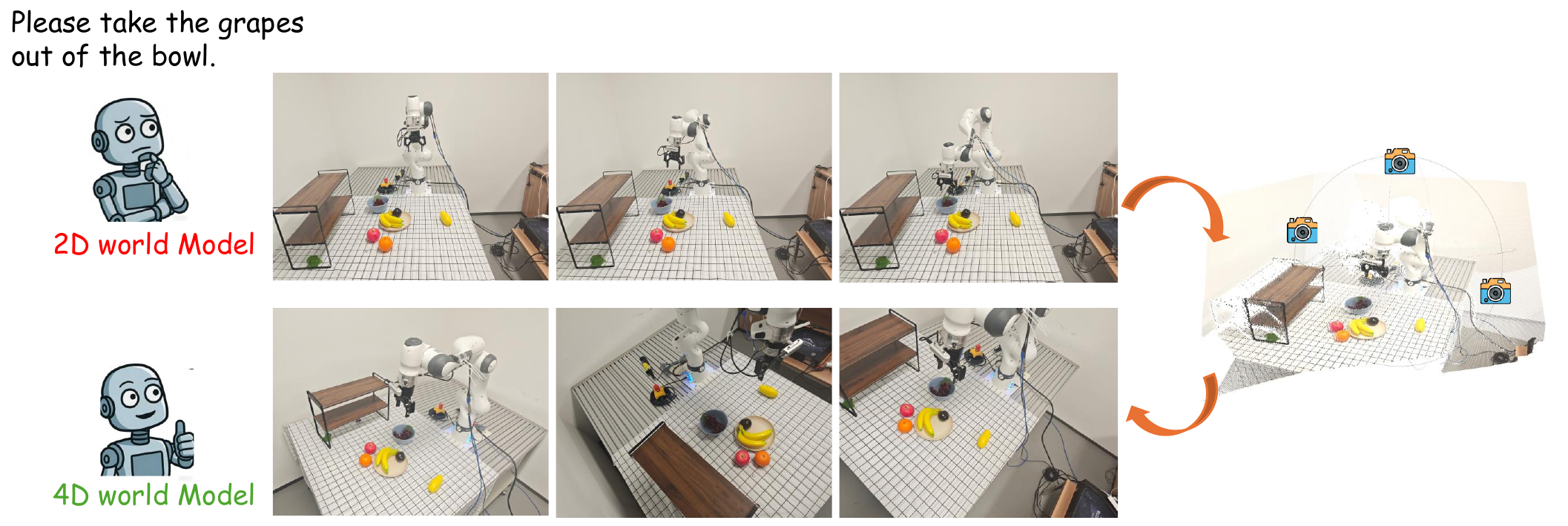}
        \caption{\textbf{Illustration of the proposed Embody4D.} A generalist data engine that transforms monocular robot videos into videos from flexible viewpoints for embodied 4D world modeling, thereby enhancing downstream task performance.}
        \vspace{-0.8cm}
        \label{fig:teaser}
    \end{figure}

    To this end, we propose \textbf{Embody4D}, a generalist data engine for embodied 4D world modeling, as illustrated in Fig.~\ref{fig:teaser}. Given a monocular robot video, Embody4D generates temporally coherent novel-view videos from flexible target viewpoints, thereby bridging the gap between limited 2D recordings and rich 4D embodied data.
    
    Specifically, 4D generative models are fundamentally constrained by the scarcity of multi-view dynamic training data, a challenge that becomes even more acute in embodied scenarios. To mitigate this issue, we first establish a \textbf{compositional 4D embodied data synthesis} pipeline to composite 30 cross-morphology manipulators from the MuJoCo Menagerie \cite{mujoco} with diverse real-world backgrounds. This heterogeneous data construction significantly enhances Embody4D’s generalization capabilities. To stabilize the generative process, we devise a \textbf{latent confidence-aware expert modulation} strategy, which converts the reliability of the warped prior into token-wise expert routing in the latent space. By adaptively assigning tokens to \textit{copy}, \textit{repair}, and \textit{inpaint} experts, our method preserves reliable projections, corrects geometric misalignments, and completes uncertain regions, thereby improving spatiotemporal coherence and texture fidelity during viewpoint transitions. To further guarantee the fidelity of complex interactions, we introduce an \textbf{interaction-aware attention} mechanism that leverages segmentation-based motion biases to decouple manipulation dynamics from the background \cite{attention}, thereby preserving the structural integrity of interacting entities. Finally, extensive experiments across diverse benchmarks demonstrate that Embody4D achieves state-of-the-art performance. Furthermore, comprehensive evaluations across simulation benchmarks and real-world robotic platforms validate our model as a robust data engine, capable of generating high-fidelity and temporally coherent video data to improve downstream robotic planning and policy learning.
    
    In summary, our key contributions can be concluded as follows:
    
    \noindent $\bullet$ We propose Embody4D, a 4D world model that synthesizes novel-view videos from monocular videos under flexible viewpoint control, providing comprehensive multi-view representations.

    \noindent $\bullet$ We design a unified generation framework with three key components: compositional data synthesis to alleviate data scarcity, latent confidence-aware expert modulation to improve cross-view consistency, and interaction-aware attention to preserve the manipulation fidelity of dynamic robot-object interactions.

    \noindent $\bullet$ We achieve state-of-the-art performance across visual benchmarks, while simulation benchmarks and real-world robotic experiments further demonstrate the effectiveness of Embody4D in empowering downstream robotic planning and policy learning.


\section{Related Works}
\textbf{Generative Video Models as Data Engines.}
Existing robotic data engines based on video generation mainly convert human object interaction videos into robot demonstrations \cite{human2robot, unidex} or synthesize robot videos conditioned on manipulator actions \cite{dreamdojo, orv, irasim, enerverse}. 
For example, Human2Robot \cite{human2robot} generates robot videos from human demonstrations and then estimates robot actions from the generated videos, but such human to robot transfer suffers from inaccurate action recovery due to the embodiment and action space gap. 
Action conditioned methods such as EnerVerse \cite{enerverse} train video world models conditioned on robot actions to generate interactive robot rollouts, yet they still struggle to strictly align generated visual dynamics with input actions and may produce physically implausible interactions, particularly in fine grained robot object manipulation. We instead explore a novel view synthesis paradigm that expands existing monocular robot videos into flexible target camera viewpoints without generating new action trajectories. By preserving accurate action annotations and enriching each trajectory with multi view observations, our approach alleviates the limited viewpoint bias of robot datasets, strengthens 3D spatiotemporal modeling and improves viewpoint generalization for downstream policy learning.

\textbf{4D Generation Model.}
Video-to-video 4D generation aims to synthesize temporally and spatially consistent dynamic 3D scenes from monocular videos, providing comprehensive environmental representations for embodied intelligence and content creation \cite{4dgeneration:Asurvey,4dworldbench,3dand4dworldmodeling}. Nevertheless, progress in this domain remains severely constrained by the scarcity of high-quality training data. \cite{Advancesin3dgeneration:Asurvey,3DSceneGeneration:ASurvey}. Early approaches \cite{sv4d} like DimensionX \cite{dimensionx} employs LoRA fine-tuning to extend single-view videos into multi-view sequences but lacks the capability for precise viewpoint control. To mitigate data and camera control limitations, recent ``warp-then-inpaint'' paradigms \cite{see4d}, represented by Trajectory Crafter \cite{trajectorycrafter}, construct pseudo-4D data with camera information from monocular videos to serve as generation conditions. Conversely, ReCamMaster \cite{recammaster} tackles these issues by synthesizing 4D data via game engines and optimizing the camera information injection mechanism. Inspired by these advancements, to address the scarcity of high-quality 4D data in embodied scenarios, we leverage the MuJoCo simulator to construct a dataset exhibiting high 4D consistency across diverse scenes and robotic arm morphologies. Furthermore, we adopt and refine the ``warp-then-inpaint'' paradigm to achieve enhanced generation quality.

\section{Methodology} 
\textbf{Preliminary: Video Diffusion Models.}
Video diffusion models generate videos by learning to reverse a gradual noising process \cite{flowmatching}. Given a clean video $x_0 \in \mathbb{R}^{n \times 3 \times h \times w}$, Gaussian noise $\epsilon \sim \mathcal{N}(0,I)$ is progressively injected to obtain noisy samples at different timesteps. To improve efficiency, latent video diffusion models perform this process in the latent space of a pretrained 3D VAE, where the video is encoded as $z_0=\mathcal{E}(x_0)$. A denoising network, commonly implemented with a Diffusion Transformer (DiT), is trained to predict the injected noise from the noisy latent $z_t$:
\begin{equation}
    \mathcal{L}_{\mathrm{diff}} =
    \mathbb{E}_{t \sim \mathcal{U}(0,1), \epsilon \sim \mathcal{N}(0,I), z_0}
    \left[
    \left\|
    \epsilon_\theta(z_t,t,c) - \epsilon
    \right\|_2^2
    \right],
\end{equation}
where $c$ denotes optional conditional information, such as text tokens, source observations, or camera conditions. During inference, Gaussian latent noise is iteratively denoised into clean latent tokens, which are decoded by the VAE decoder $\mathcal{D}$ to produce the final video $\hat{x}=\mathcal{D}(z_0)$ \cite{diffusion}.

\begin{figure}[t]
    \raggedright
    \includegraphics[width=1\linewidth]{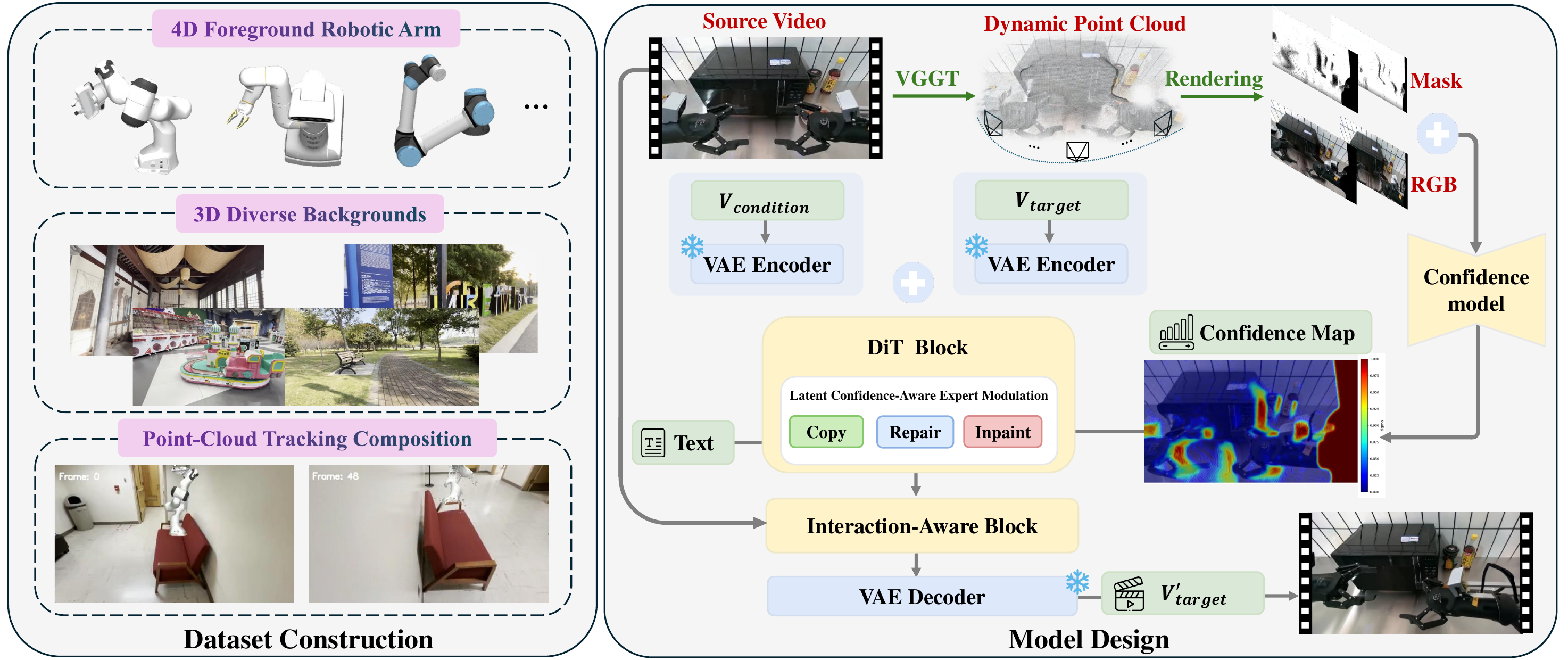} 
    \caption{\textbf{Overview of Embody4D.} We build paired embodied 4D training data through compositional synthesis and adopt a “warp-then-inpaint” framework for novel-view video generation. Given a source video, Embody4D reconstructs dynamic geometry, warps RGB/masks to target views, and generates the final video with confidence-aware modulation and interaction-aware attention.}
    \vspace{-0.5cm}
    \label{fig:pipeline}
\end{figure}

\textbf{Compositional 4D Embodied Data Synthesis.}
To address the scarcity of embodied 4D data and improve cross-embodiment generalization, we design a compositional data synthesis strategy for constructing view-consistent robotic 4D videos. As shown in Fig. \ref{fig:pipeline} (left), we select 30 embodied models from MuJoCo Menagerie \cite{mujoco_m}, including humanoid robots, single-arm manipulators, dual-arm manipulators, and small grippers. Each model is controlled in MuJoCo to perform random motions around its default pose, producing diverse foreground robot dynamics. For background construction, we sample image pairs from DL3DV \cite{dl3dv}. Candidate pairs are first selected by minimizing camera translation within fixed-interval frames, and then filtered with GPT-4o \cite{gpt} to ensure that the scene center remains visible across views, resulting in temporally consistent real-world backgrounds.

To obtain view-consistent foreground-background composition, we configure two virtual MuJoCo cameras whose extrinsics are aligned with the selected DL3DV image pair. The robot is first composited into the reference view with random rotation and scaling. For the second view, we further introduce a 3D anchor tracking strategy to preserve geometric consistency across viewpoints. Specifically, VGGT \cite{vggt} is used to reconstruct the 3D geometry of source-view video and estimate per-pixel depth together with camera poses. Let $\mathcal{I}_{src}$ denote the first 49 frames of the source video, we randomly sample a subset $\mathcal{S} \subset \mathcal{I}_{src}$ with $|\mathcal{S}|=5$. For each sampled frame $k \in \mathcal{S}$, we lift the robot center $\mathbf{u}_k$ to the VGGT reconstructed world coordinate system and then transform it into the target camera coordinate system as
\begin{equation}
    \mathbf{P}^{c}_{k \to t'} =
    \mathbf{R}_{t'}
    \mathbf{R}^{-1}_{k}
    \left(
        z_k \mathbf{K}^{-1}\tilde{\mathbf{u}}_k - \mathbf{T}_{k}
    \right)
    + \mathbf{T}_{t'},
    \label{eq:target_camera_anchor}
\end{equation}
where $z_k$ is the estimated depth, $\tilde{\mathbf{u}}_k$ is the homogeneous image coordinate, $(\mathbf{R}_{k}, \mathbf{T}_{k})$ denotes the source-view extrinsics, and $(\mathbf{R}_{t'}, \mathbf{T}_{t'})$ denotes the VGGT-estimated target-view extrinsics. To reduce depth and pose noise, we average the transformed points in the target camera coordinate system to obtain a robust 3D anchor:
\begin{equation}
    \bar{\mathbf{P}}^{c}_{t'} =
    \frac{1}{|\mathcal{S}|}
    \sum_{k \in \mathcal{S}}
    \mathbf{P}^{c}_{k \to t'}.
    \label{eq:anchor_averaging}
\end{equation}
Finally, this anchor is projected onto the target image plane as $\tilde{\mathbf{u}}_{t'} \sim \mathbf{K}\bar{\mathbf{P}}^{c}_{t'}$, followed by homogeneous normalization to obtain the target-view robot center $\mathbf{u}_{t'}$ for accurate second-view composition.

\textbf{Latent Confidence-Aware Expert Modulation.}
Recent 4D novel-view generation methods often follow a ``warp-then-inpaint''
paradigm \cite{trajectorycrafter, see4d}, where source-view observations are first projected to the target
camera using reconstructed geometry and then refined by a generative model.
However, the projected prior is spatially heterogeneous: well-aligned regions
should be preserved, geometrically inconsistent regions require correction, and
disoccluded or highly uncertain areas demand completion. This motivates a
region-adaptive formulation of target-view synthesis, rather than uniform
denoising. 
To this end, we propose Latent Confidence-Aware Expert Modulation, which decomposes target-view generation into three confidence-conditioned behaviors:
\textit{copy}, \textit{repair}, and \textit{inpaint} \cite{gta}. Instead of relying on a
binary valid mask to impose a coarse projected/unprojected partition, we learn a
continuous latent confidence map that estimates the reliability of the warped
prior. This map provides a soft routing signal: reliable regions are encouraged
to copy the warped content, moderately unreliable regions are routed to repair,
and highly uncertain or missing regions are routed to inpaint.

To supervise this routing signal, we use the warp-to-target discrepancy in the
VAE latent space as an oracle measure of correction demand \cite{inversecrafter}. Given a warped video
$\mathbf{x}_{w}$ and its ground-truth target video $\mathbf{x}_{t}$, a frozen
VAE encoder $\mathcal{E}(\cdot)$ \cite{cogvideo} maps them into a shared latent space. The
latent residual
$\mathcal{E}(\mathbf{x}_{t})-\mathcal{E}(\mathbf{x}_{w})$ characterizes the
ideal correction required to transform the warped prior toward the target. Its
magnitude naturally reflects the reliability of the warped latent: smaller
discrepancies indicate content that can be largely preserved, while larger
discrepancies reveal regions that require repair or completion.

To obtain a routing signal without relying on the ground-truth target video, we
train a lightweight 3D U-Net $f_{\phi}$ \cite{unet} to predict the latent correction from
available conditioning inputs, including the warped latent $\mathcal{E}(\mathbf{x}_{w})$ and the geometric
mask $\mathbf{M}_{s}$ resized to the latent resolution:
\setlength{\abovedisplayskip}{4pt}
\setlength{\belowdisplayskip}{4pt}
\begin{equation}
    \mathcal{L}_{\Delta}
    =
    \left\|
    f_{\phi}
    \left(
    \mathcal{E}(\mathbf{x}_{w}),
    \mathrm{Resize}(\mathbf{M}_{s})
    \right)
    -
    \left(
    \mathcal{E}(\mathbf{x}_{t})
    -
    \mathcal{E}(\mathbf{x}_{w})
    \right)
    \right\|_2^2 .
\end{equation}
We aggregate the predicted residual magnitude along the channel dimension to
obtain a scalar correction-demand map, which is then converted into a continuous
routing cue for the \textit{copy}, \textit{repair}, and \textit{inpaint}
experts.

During generative training, this routing cue is tokenized and mapped into
expert weights
$\mathbf{A}^{\mathrm{tok}}\in\mathbb{R}^{B\times N\times 3}$. For the $l$-th
Transformer block, let
$\mathbf{u}_{n}^{l}=\mathrm{FFN}^{l}(\bar{\mathbf{h}}_{n}^{l})$ denote the
standard FFN output of token $n$, where $\bar{\mathbf{h}}_{n}^{l}$ is the
normalized hidden state before the FFN. We augment this FFN branch with a
routing-adaptive expert residual:
{
\setlength{\abovedisplayskip}{4pt}
\setlength{\belowdisplayskip}{4pt}
\begin{equation}
    \mathbf{h}_{n}^{l+1}
    =
    \mathbf{h}_{n}^{l}
    +
    \mathbf{g}_{n}^{l}
    \odot
    \left[
    \mathbf{u}_{n}^{l}
    +
    s
    \sum_{r \in \{\mathrm{copy},\mathrm{repair},\mathrm{inpaint}\}}
    \mathbf{A}_{n,r}^{\mathrm{tok}}
    E_{r}^{l}
    \left(
    \mathbf{u}_{n}^{l}
    \right)
    \right].
\end{equation}
}

Here, $E_r^l$ denotes the expert for route $r$,
$\mathbf{A}_{n,r}^{\mathrm{tok}}$ is the corresponding token-level routing
weight, $s$ controls the expert residual strength, and $\mathbf{g}_{n}^{l}$ is
the FFN gate inherited from the Transformer block.This design preserves the original FFN pathway as a stable generative backbone,
while using confidence-aware expert residuals to adaptively allocate modeling
capacity across copied, repaired, and inpainted regions.

\begin{figure}[t]
    \centering
    \includegraphics[width=0.9\linewidth]{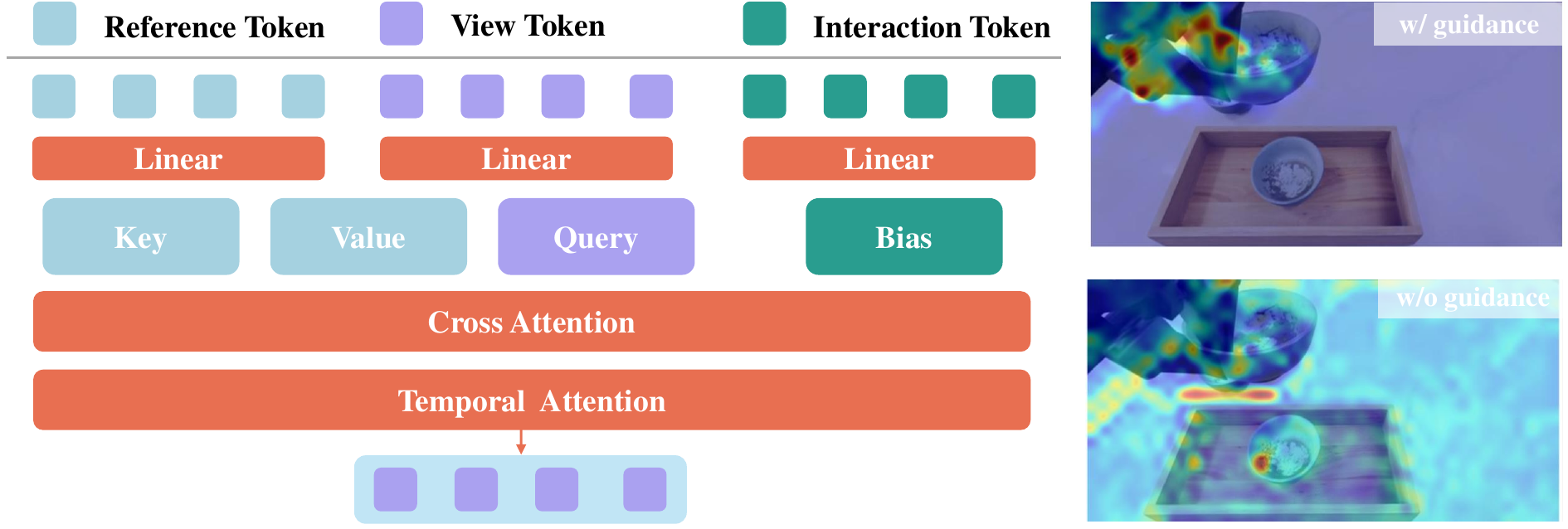}
    \caption{\textbf{Interaction-Aware Block.} The module injects mask-derived interaction biases into cross attention to prioritize manipulation regions and enhance cross-view consistency. (qualitative comparison in heatmaps).}
    \vspace{-0.7cm}
    \label{fig:interaction}
\end{figure}

\textbf{Embodied Interaction-Aware Attention.}
The core idea of Embodied Interaction-Aware Attention is to explicitly bias the model toward dynamic robot-object interaction regions during novel-view synthesis. 
To compensate for the limited object manipulation in our synthetic data, we
construct pseudo-4D pairs from monocular robotic manipulation videos using a
forward-backward view cycle: the source video is warped to a sampled novel view
and then warped back to the original source view, yielding pseudo paired
supervision for view-consistent embodied 4D generation \cite{trajectorycrafter}.
For these data, we extract temporally consistent interaction masks $\mathbf{M}$ that cover both the robotic arm and manipulated objects. 
Specifically, we adopt MemFlow \cite{memflow} for simulation or stable videos, and use SAM3 \cite{sam3} for real-world sequences with camera jitter. 
For our synthetic data, the arm foreground masks are directly obtained from rendering. 
These masks provide explicit motion priors, indicating where the model should focus when synthesizing high-dynamic embodied interactions.

Given queries $\mathbf{Q}$ from the synthesized features and keys/values $\mathbf{K}, \mathbf{V}$ from the reference features \cite{ediff}, we decompose attention into a global path and an interaction-guided path:
\begin{equation}
\begin{array}{c}
\displaystyle
\mathbf{O}_{\mathrm{global}}
=
\operatorname{Softmax}
\left(
\frac{\mathbf{Q}\mathbf{K}^{\top}}{\sqrt{d_k}}
\right)
\mathbf{V},
\\[0.8em]
\displaystyle
\mathbf{O}_{\mathrm{guided}}
=
\operatorname{Softmax}
\left(
\frac{\mathbf{Q}\mathbf{K}^{\top}}{\sqrt{d_k}} + \mathbf{B}
\right)
\mathbf{V},
\quad
\mathbf{B}_{i,j}
=
\begin{cases}
\lambda, & \mathbf{M}_{i,j}=1,\\
0, & \text{otherwise}.
\end{cases}
\end{array}
\end{equation}
Here, $\mathbf{M}_{i,j}$ denotes a pairwise token-level interaction indicator, where $\mathbf{M}_{i,j}=1$ if both query token $i$ and key token $j$ correspond to the robot-object interaction region after downsampling the pixel-level mask to the attention resolution, and $\mathbf{M}_{i,j}=0$ otherwise. The global path preserves overall scene context, while the guided path increases the attention response on interaction regions through the mask-induced bias $\mathbf{B}$.
Finally, the two paths are fused with a curriculum coefficient $\alpha$:
\begin{equation}
    \mathbf{O} = (1-\alpha)\mathbf{O}_{\text{global}} + \alpha \mathbf{O}_{\text{guided}},
\end{equation}
allowing the model to progressively shift from global structure modeling to fine-grained embodied interaction modeling during training.

\section{Experiments}
\textbf{Dataset.} 
To train this model, we initially utilize a curated dataset of 23K 4D samples synthesized via foreground-background composition, enabling the model to learn diverse robotic arm morphologies and enhancing its 4D consistency. Subsequently, we leverage 24K monocular embodied data uniformly sampled from five datasets (AGIBOT \cite{agibot}, Rh20t \cite{rh20t}, Robset \cite{roboagent}, Bc-z \cite{bcz}, and Interndata-A1 \cite{interndata}) to learn real-world robotic arm interactive operations.

We evaluate our method on a held-out test set of 120 monocular videos that are disjoint from the training data and cover diverse robotic embodiments, scenes, and manipulation scenarios.

\textbf{Implementation.}
We fine-tune Embody4d based on the pretrained TrajectoryCrafter \cite{trajectorycrafter} architecture. During training, the frame resolution is fixed at 384×672, and the video length is set to 49 frames. Both training stages use a batch size of 2 and are conducted on 8 A100 GPUs. We apply vggt \cite{vggt} as the reconstruction foundation model for estimating camera parameters and depth maps. 

\textbf{Baselines and Metrics.}
For a fair comparison, we evaluate our approach against several state-of-the-art baselines, including camera-controlled video generation models and point cloud rendered multi-view generation models, i.e., TrajectoryCrafter \cite{trajectorycrafter}, ReCamMaster \cite{recammaster}, Ex-4D \cite{ex4d}, and Reangle-A-Video \cite{reangle}. For video generation evaluation, we strictly follow the VBench \cite{vbench} protocol, reporting scores across multiple dimensions: Subject Consistency, Background Consistency, Temporal Flickering, Motion Smoothness, and Imaging Quality. In addition, we evaluated the 3D consistency metric on the MEt3R \cite{met3r} benchmark. 
\subsection{Comparison With Other Methods}
We compare our method with existing baselines in Fig. \ref{fig:compare} and Tab. \ref{tab:results}. Qualitatively, ReCamMaster suffers from artifacts and geometric distortions under large viewpoint changes, as its view transformation is mainly driven by camera extrinsics without exploiting source-video geometric priors. Reangle-A-Video and Ex-4D, which rely on LoRA-based fine-tuning, often exhibit incomplete hole filling and degraded visual quality in challenging regions, while the previous state-of-the-art TrajectoryCrafter tends to hallucinate fine structures around robotic arms and grippers. In contrast, our method produces high-fidelity novel-view embodied videos with stronger 4D consistency and more reliable manipulation details. Quantitatively, our approach achieves the best results across all evaluated metrics, establishing a new state of the art and demonstrating superior robustness, visual realism, and temporal stability for flexible-view embodied video synthesis. More qualitative comparisons are provided in Fig. \ref{fig:result3}, \ref{fig:result4}

\begin{table}[t]
\centering
\scriptsize 
\caption{\textbf{Quantitative comparisons.} \textbf{Bold}: Best. \underline{Underline}: Second Best. Our method consistently outperforms previous solutions on both VBench and MEt3R}
\label{tab:results}
\begin{tabular}{l||ccccc|c} 
\toprule 
 & \multicolumn{5}{c}{\textbf{VBench} $\uparrow$} & \textbf{MEt3R} $\downarrow$ \\
\textbf{Method} & Subject & Background & Temporal & Motion & Imaging & 3D Consistency \\ 
\midrule 
ReCamMaster \cite{recammaster}         & 0.8981 & 0.8976 & \underline{0.9717} & 0.9841 & 0.5914 & 0.2454 \\
Ex-4D \cite{ex4d}               & 0.8088 & 0.8906 & 0.9213 & 0.9742 & 0.5732 & 0.2713 \\
Reangle-A-Video \cite{reangle}        & 0.9152 & 0.9224 & 0.9711 & 0.9879 & \underline{0.6437} & 0.2288 \\
TrajectoryCrafter \cite{trajectorycrafter}     & \underline{0.9202} & \underline{0.9388} & 0.9714 & \underline{0.9911} & 0.6257 & \underline{0.2040} \\
\textbf{Ours}        & \textbf{0.9351} & \textbf{0.9491} & \textbf{0.9734} & \textbf{0.9937} & \textbf{ 0.6566} & \textbf{0.1681} \\
\bottomrule
\end{tabular}
\end{table}

\begin{figure}[t]
    \raggedright
    \vspace{-0.4cm}
    \includegraphics[width=1\linewidth]{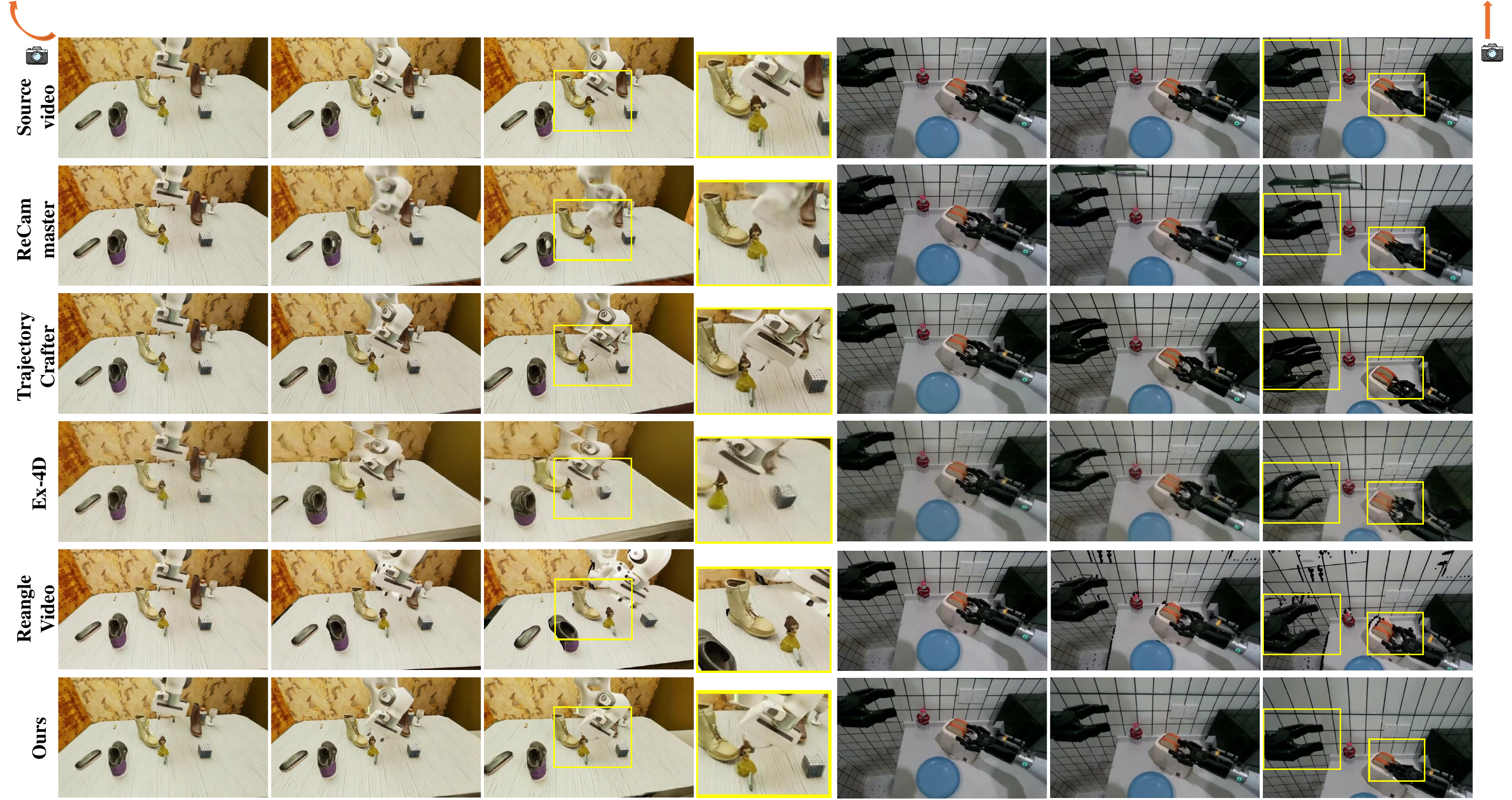}
    \caption{\textbf{Qualitative comparisons of novel view video synthesis.} Our method achieves better structural consistency and visual quality than baselines.}
    \vspace{-0.5cm}
    \label{fig:compare}
\end{figure}

\subsection{Ablation Study}
\begin{wraptable}{r}{0.50\columnwidth}
\vspace{-1.0em}
\centering
\scriptsize
\setlength{\tabcolsep}{3pt}
\caption{Ablation study on video generation quality metrics.}
\label{tab:ablation}
\begin{tabular}{l||ccccc}
\toprule
\textbf{Method} 
& PSNR $\uparrow$ 
& SSIM $\uparrow$ 
& LPIPS $\downarrow$ 
& MEt3R $\downarrow$ \\ 
\midrule
Baseline & 19.0293 & 0.6564 & 0.3303 & 0.1923 \\
+ data & 23.1277 & 0.7899 & 0.2136 & 0.1757 \\
+ data + IA & 23.4920 & 0.7907 & 0.2112 & 0.1702 \\
\textbf{+ data + IA + Expert} 
& \textbf{23.6397} 
& \textbf{0.8064} 
& \textbf{0.1846} 
& \textbf{0.1681} \\
\bottomrule
\end{tabular}
\vspace{-1.0em}
\end{wraptable}
We conduct ablation studies to quantify the contribution of each component. Table~\ref{tab:ablation} reports the results of our framework. The baseline is the TrajectoryCrafter backbone fine-tuned only on embodied pseudo-4D data. Adding our compositional 4D embodied data (\textit{+data}) and interaction-aware attention (\textit{+data+IA}) progressively improves generation quality. The full model (\textit{+data+IA+Expert}), further equipped with latent confidence-aware expert modulation, achieves the best performance, improving PSNR by 24.23\% and SSIM by 22.85\%, while reducing LPIPS by 44.11\% and MEt3R by 12.58\%. These results validate the effectiveness of our data curation, interaction-aware attention, and confidence-aware modulation for stable embodied 4D generation. As illustrated in Fig.~\ref{fig:ab} and Fig.~\ref{fig:abdata}, both the proposed modules and the curated robotic data enhance temporal stability, interaction fidelity, and cross-embodiment generalization.

\subsection{Downstream Task Performance Evaluation}
\textbf{Simulation Experiments.}
In our simulation experiments, we used LIBERO-Plus~\cite{fei25libero-plus}, a general-purpose single-arm manipulation benchmark built on LIBERO~\cite{liu2023libero}, and introduced seven perturbation settings to evaluate policy robustness. To ensure a fair comparison, we fine-tuned $\pi_{0.5}$~\cite{0.5} on the LIBERO dataset augmented using our method and compared it with the same $\pi_{0.5}$ model fine-tuned on the original LIBERO dataset. Figure~\ref{fig:libero} presents representative generation results, while Figure~\ref{fig:compare} shows an example comparison.



\textbf{Real World Experiments.}
Embodied world models offer a promising data engine for robot learning~\cite{enerverse, embodiedgen}. We study whether Embody4D can enhance multi-view perception for real-world deployment by augmenting monocular robot data with synthesized novel views. Using Embody4D, ReCamMaster, and TrajectoryCrafter, we generate an additional third-person view from a single external-camera dataset and train the state-of-the-art VLA policy $\pi_{0.5}$~\cite{0.5} on the augmented data.

\begin{wrapfigure}{r}{0.55\linewidth}
    \vspace{-10pt}
    \centering
    \includegraphics[width=\linewidth]{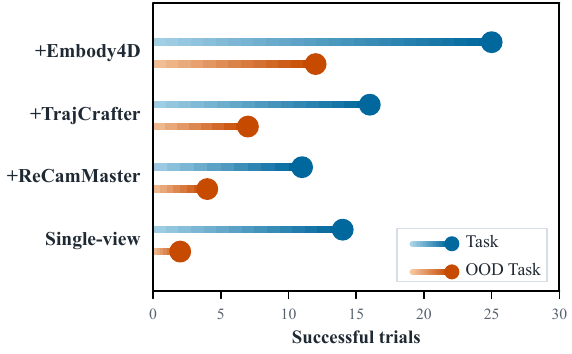}
    \caption{\textbf{real-world policy success counts} under four experimental settings on seen tasks and OOD unseen tasks}
    \vspace{-10pt}
    \label{fig:task}
\end{wrapfigure}

We evaluate on real-world tabletop manipulation with a Franka Research 3 (FR3) arm \cite{franka} and a Robotiq 2F-85 gripper. The setup contains two fixed external cameras and one wrist-mounted camera. From 100 human demonstrations, we train four variants: a single-view baseline, Embody4D-augmented two-view training, ReCamMaster-augmented training, and TrajectoryCrafter-augmented training. Policies are tested on five pick-and-place tasks with randomized object placements, including three seen tasks and two unseen OOD tasks, with 10 trials per task.

As shown in Tab.~\ref{tab:realworld} and Fig.~\ref{fig:task}, Embody4D achieves the highest success rate of 74\% and shows superior OOD generalization. These results demonstrate that Embody4D can effectively function as a data engine for downstream embodied policy learning.

\begin{table}[t]
\centering
\scriptsize
\renewcommand{\arraystretch}{0.80}
\setlength{\tabcolsep}{2.2pt}
\caption{Real-world embodied experiments on FR3 with $\pi_{0.5}$ Droid. Each task has 10 trials with randomized object positions. T1--T3 are seen tasks and T4--T5 are unseen tasks (OOD).}
\label{tab:realworld}
\begin{tabular}{l||ccc|c}
\toprule
\textbf{Task} 
& \textbf{Single-view} 
& \textbf{ReCamMaster \cite{recammaster}} 
& \textbf{TrajCrafter \cite{trajectorycrafter}} 
& \textbf{Embody4D} \\
\midrule
T1 (Grapes$\rightarrow$Bowl)   & 5/10 & 4/10 & 6/10 & \textbf{8/10} \\
T2 (Grapes$\rightarrow$Plate)  & 5/10 & 5/10 & 6/10 & \textbf{8/10} \\
T3 (Mangoes$\rightarrow$Bowl)  & 4/10 & 2/10 & 4/10 & \textbf{9/10} \\
T4 (Lemons$\rightarrow$Bowl)   & 1/10 & 2/10 & 0/10 & \textbf{6/10} \\
T5 (Bananas$\rightarrow$Plate) & 1/10 & 2/10 & 7/10 & \textbf{6/10} \\
\midrule
SR & 32\% & 30\% & 46\% & \textbf{74\%} \\
\bottomrule
\end{tabular}
\vspace{-4mm}
\end{table}
\section{Limitation}
A limitation of our current framework lies in the limited temporal horizon of existing video generation models, which commonly support only fixed-length generation, such as 49 or 81 frames \cite{wan, cogvideo, cosmos}. As a result, long input videos must be split into multiple clips for novel-view synthesis. Since each clip is generated independently, stitching them into a full sequence may introduce temporal discontinuities at clip boundaries, causing mild jitter and slightly degrading visual quality. Additional failure cases are shown in Fig.~\ref{fig:failure}.
\section{Conclusion}
We presented Embody4D, a video-to-video 4D world model for Embodied AI that synthesizes flexible novel views from monocular videos. Embody4D combines a scalable compositional data synthesis pipeline with confidence-aware expert modulation and interaction-aware attention, enabling geometrically consistent, temporally stable, and interaction-faithful generation. Extensive experiments demonstrate state-of-the-art synthesis quality and show that Embody4D serves as an effective data engine for downstream robotic policy learning in both simulation and real-world deployment.

\clearpage

\bibliography{example}  
\newpage
\appendix

\section{Ablation Study}
\begin{figure}[h]
    \raggedright
    \includegraphics[width=1\linewidth]{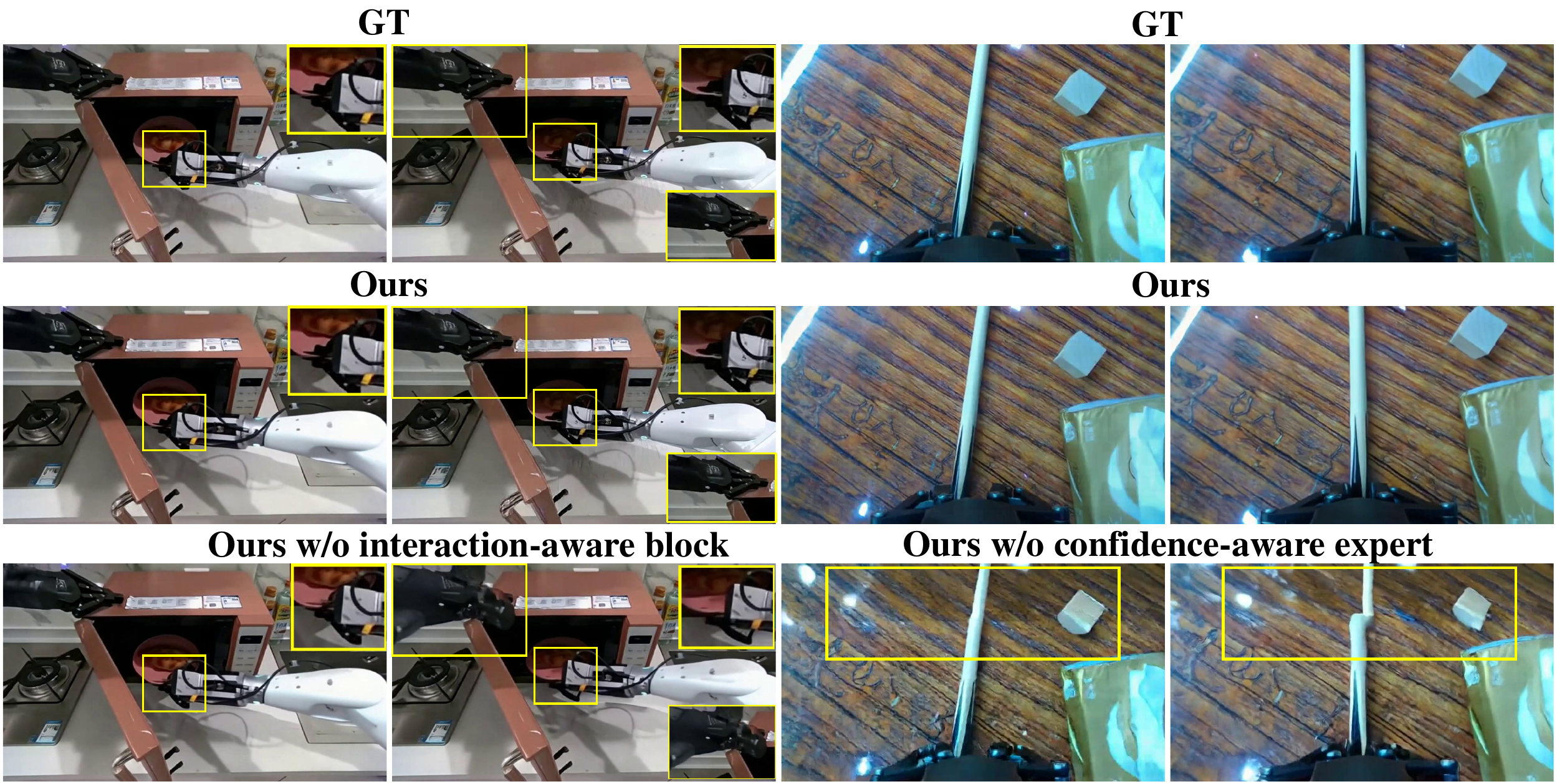}
    \caption{\textbf{Qualitative comparisons of the ablations.} Results of ablation on embodied interaction-aware attention and confidence-aware expert modulation strategy.}
    \label{fig:ab}
    \vspace{-0.4cm}
\end{figure}

\textbf{Ablations of Embody4d.} To systematically evaluate the contribution of each module within Embody4D, we conducted ablation studies on two core architectural components: the latent confidence-aware expert modulation mechanism and the interaction-aware attention module. As illustrated in Fig. \ref{fig:ab}, the confidence-aware strategy performs reliability-conditioned expert routing, adaptively assigning spatial regions to the \textit{copy}, \textit{repair}, and \textit{inpaint} experts based on their estimated projection reliability. This approach preserves fine-grained structural details in high-confidence areas while ensuring seamless geometric and textural transitions, thereby significantly enhancing the spatio-temporal consistency and visual quality of the synthesized videos. Furthermore, the interaction-aware attention mechanism introduces a spatial bias focused on manipulation zones, directing the model to prioritize critical robot-object interactions. This ensures superior fidelity in delicate manipulation regions and strengthens the physical plausibility of complex scenarios. Quantitative results in Tab. \ref{tab:ablation} further demonstrate that these two modules synergistically bolster the stability of Embody4D, enabling the synthesis of highly realistic 4D robotic interaction videos with consistent multi-view structures.

\vspace{-0.3cm}

\begin{figure}[t]
    \raggedright
    \includegraphics[width=1\linewidth]{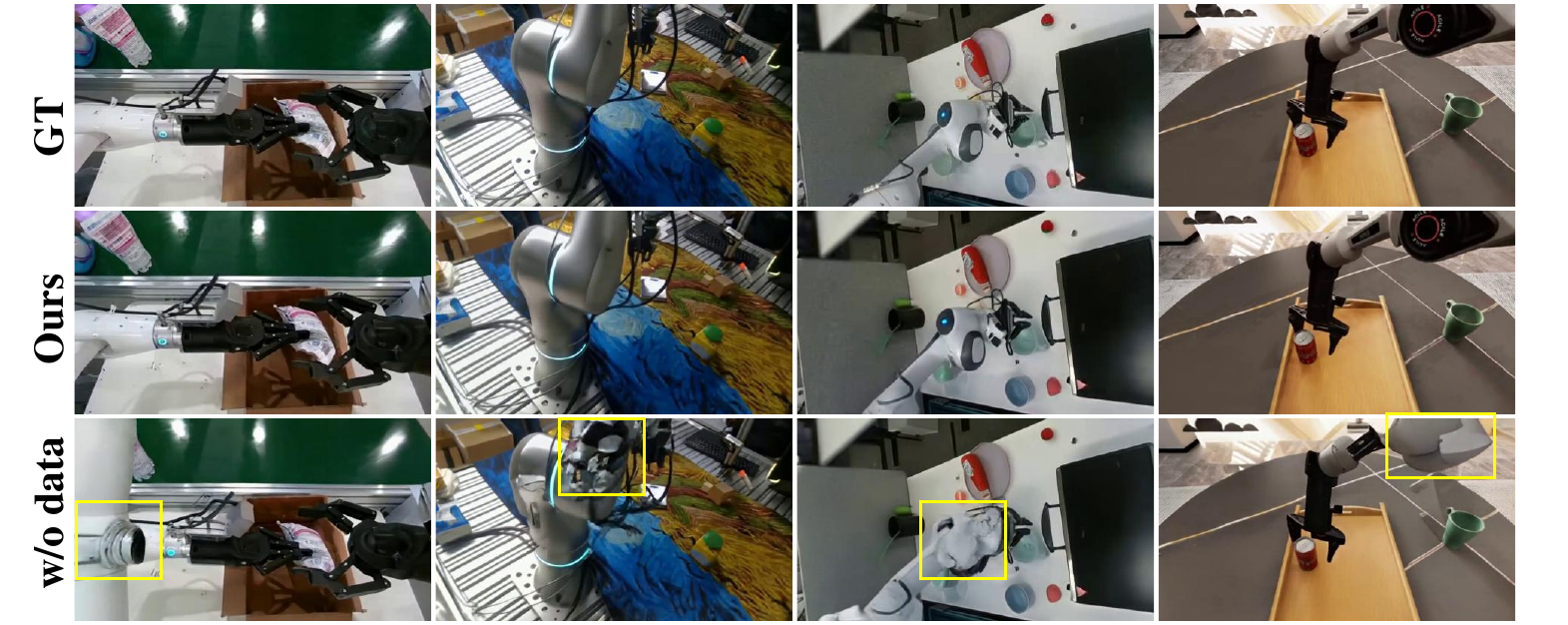}
    \caption{\textbf{Ablation on the compositional training data.} Comparison between our full model and a variant trained without compositional data. The yellow boxes show that our strategy is key to mitigating geometric distortions and ensuring generalization across diverse robotic morphologies.}
    \label{fig:abdata}
\end{figure}

\textbf{Ablation on the training data.}
The dataset curation strategy represents another key contribution of our work, where the model is trained on a hybrid dataset comprising open-source monocular embodied data and compositionally constructed diverse robotic embodied data. To validate its effectiveness, we conducted an ablation study comparing models trained with and without the compositionally constructed data. Qualitative results in Fig. \ref{fig:abdata} demonstrate that the model lacking diverse robotic data struggles to generalize to unseen robotic structures, frequently producing implausible geometric hallucinations in occluded regions. Conversely, our hybrid training strategy enables the model to internalize robust structural priors, allowing it to synthesize physically consistent robotic poses across various morphologies while maintaining a motion distribution that aligns with the source video. The quantitative results in Tab. \ref{tab:ablation} further substantiate that this curated diversity is essential for achieving strong generalization in complex manipulation scenarios.

\begin{figure}[t]
    \raggedright
    \includegraphics[width=1\linewidth]{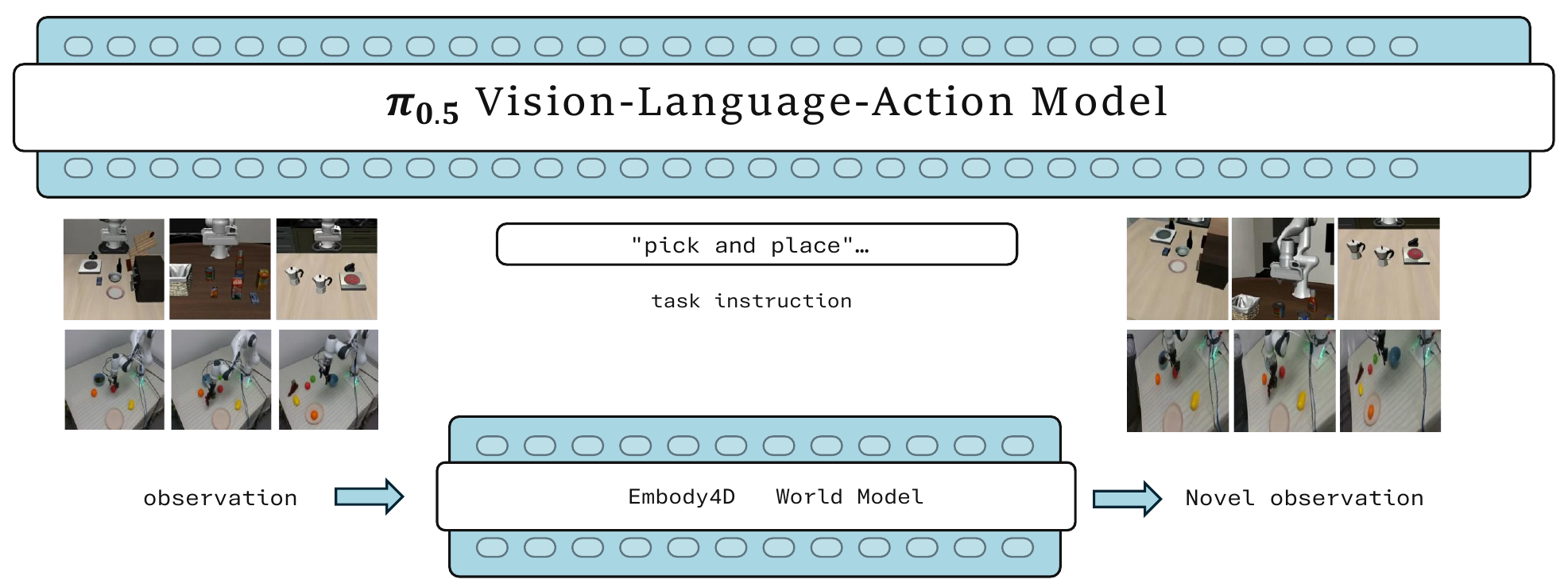}
    \caption{\textbf{Illustration of downstream task evaluation.}
Embody4D generates flexible novel views from a single input viewpoint, providing more complete geometric information to augment the training of $\pi_{0.5}$ \cite{0.5}.}
    \label{fig:vla}
    \vspace{-0.5cm}
\end{figure}

\section{Downstream Task Performance Evaluation}
\subsection{Simulation Experiments}

\begin{table}[t]
\centering
\caption{\textbf{Effect of Embody4D augmentation on LIBERO.}
We evaluate $\pi_{0.5}$ on the LIBERO benchmark and compare training with the original LIBERO data, Embody4D-generated views, and matched real multi-view data.}
\label{tab:libero}
\resizebox{\linewidth}{!}{
\begin{tabular}{lccccc}
\toprule
\multirow{2}{*}{Training Setting} 
& \multicolumn{5}{c}{LIBERO} \\
\cmidrule(lr){2-6}
& LIBERO-10 & LIBERO-Goal & LIBERO-Spatial & LIBERO-Object & Total \\
\midrule
$\pi_{0.5}$ \cite{0.5}
& \cellcolor{green!12}\textbf{92.4\%} 
& \cellcolor{green!12}\textbf{98.0\%} 
& 98.8\% 
& 98.2\% 
& \cellcolor{green!12}\textbf{96.9\%} \\

$\pi_{0.5}$ + Embody4D 
& 90.6\% 
& 95.8\% 
& 98.2\% 
& \cellcolor{green!12}\textbf{98.6\%}
& 95.8\% \\

$\pi_{0.5}$ + Real Multi-view Data
& 91.2\% 
& 96.0\% 
& \cellcolor{green!12}\textbf{99.0\%} 
& 98.2\% 
& 96.1\% \\
\bottomrule
\end{tabular}
}
\end{table}

\begin{table}[t]
\centering
\caption{\textbf{Effect of Embody4D augmentation on LIBERO-Plus.}
We evaluate $\pi_{0.5}$ under seven LIBERO-Plus perturbation settings and compare training on the original LIBERO data with training using Embody4D-generated random-view samples and matched real multi-view data.}
\label{tab:libero_plus}
\resizebox{\linewidth}{!}{
\begin{tabular}{lcccccccc}
\toprule
\multirow{2}{*}{Training Setting} 
& \multicolumn{8}{c}{LIBERO-Plus} \\
\cmidrule(lr){2-9}
& Camera & Robot & Language & Light & Background & Noise & Layout & Total \\
\midrule
$\pi_{0.5} \cite{0.5}$ 
& 64.8\% & 71.8\% & 83.0\% & 93.5\% & 92.2\% & 78.8\% & 85.5\% & 81.4\% \\

$\pi_{0.5}$ + Embody4D 
& \cellcolor{green!12}\textbf{86.8\%} 
& 50.7\% 
& 73.2\% 
& 86.3\% 
& 82.5\% 
& \cellcolor{green!12}\textbf{86.1\%} 
& 76.5\% 
& 77.0\% \\

$\pi_{0.5}$ + Real Multi-view Data
& 81.3\% & 55.9\% & 75.8\% & 87.4\% & 78.9\% & 85.4\% & 76.9\% & 77.0\% \\
\bottomrule
\end{tabular}
}
\end{table}
For simulation experiments, we use Embody4D to generate view-randomized training videos, while evaluating policies from a fixed third-person camera. This setting isolates the effect of synthetic view augmentation on cross-view generalization and policy robustness.

For real-world experiments, we introduce an additional Embody4D-generated third-person view during training and inference. By expanding the observational coverage of robot-object interactions, this setting evaluates whether Embody4D can improve spatial reasoning, interaction understanding, and manipulation performance.

\subsection{Simulation Experiments}
We first evaluate Embody4D on the LIBERO benchmark \cite{fei25libero-plus} under three training settings:
(i) $\pi_{0.5}$ trained on the original LIBERO dataset;
(ii) $\pi_{0.5}$ fine-tuned with additional random-view samples generated by Embody4D;
and (iii) $\pi_{0.5}$ fine-tuned with real multi-view observations captured at the same target viewpoints, which provides an oracle reference for viewpoint augmentation.
For random-view generation, we fix the camera to look at the LIBERO world \cite{liu2023libero} origin and randomly sample one of six perturbation directions: up, down, left, right, forward, and backward.
The left/right rotations are sampled within $20^\circ$--$40^\circ$, the up/down rotations within $10^\circ$--$20^\circ$, and the forward/backward translations within $10$--$20\,\mathrm{cm}$.

As shown in Table~\ref{tab:libero}, all three training settings achieve strong performance on the original LIBERO benchmark, with only marginal performance gaps.
This indicates that the standard LIBERO benchmark is relatively saturated under these settings and may not fully reveal the effect of viewpoint augmentation.
We therefore further evaluate the three trained policies on the more challenging LIBERO-Plus benchmark, as reported in Table~\ref{tab:libero_plus}.

The results show that training with Embody4D-generated random-view samples achieves performance close to the real multi-view oracle, demonstrating that Embody4D preserves sufficient visual fidelity and task-relevant spatial information for downstream policy learning.
Notably, under the camera perturbation setting, Embody4D improves the success rate from $64.8\%$ to $86.8\%$, yielding a \textbf{$+22.0\%$} improvement. Under the noise perturbation setting, Embody4D further improves the success rate from $78.8\%$ to $86.1\%$.

Interestingly, Embody4D slightly outperforms the real multi-view oracle on camera, noise, and background perturbations.
We attribute this to the mild background variation and small camera jitter introduced during the novel-view generation process, which act as implicit data augmentation and improve policy robustness beyond the clean real-view setting.
For the remaining perturbation types, Embody4D shows a moderate drop compared with the original $\pi_{0.5}$ baseline.
This is expected because our augmentation primarily introduces viewpoint variations, whereas LIBERO-Plus additionally includes non-viewpoint shifts such as language, light, and layout changes.

Overall, the simulation results demonstrate that Embody4D-generated views provide effective multi-view augmentation for downstream policy learning, achieving near-oracle performance and substantially improving policy generalization to viewpoint changes.

\begin{figure}[t]
    \raggedright
    \includegraphics[width=1\linewidth]{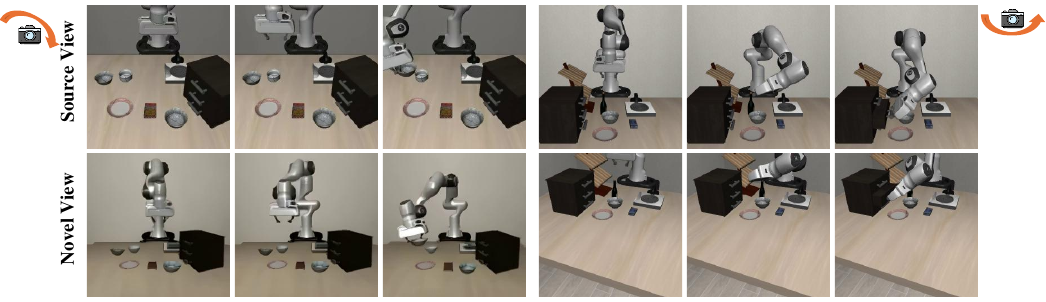}
    \caption{\textbf{Qualitative results on the LIBERO benchmark.} Our method synthesizes plausible videos from randomly varying viewpoints while preserving visual fidelity and temporal consistency.}
    \label{fig:libero}
    \vspace{-0.3cm}
\end{figure}

\begin{figure}[t]
    \raggedright
    \includegraphics[width=1\linewidth]{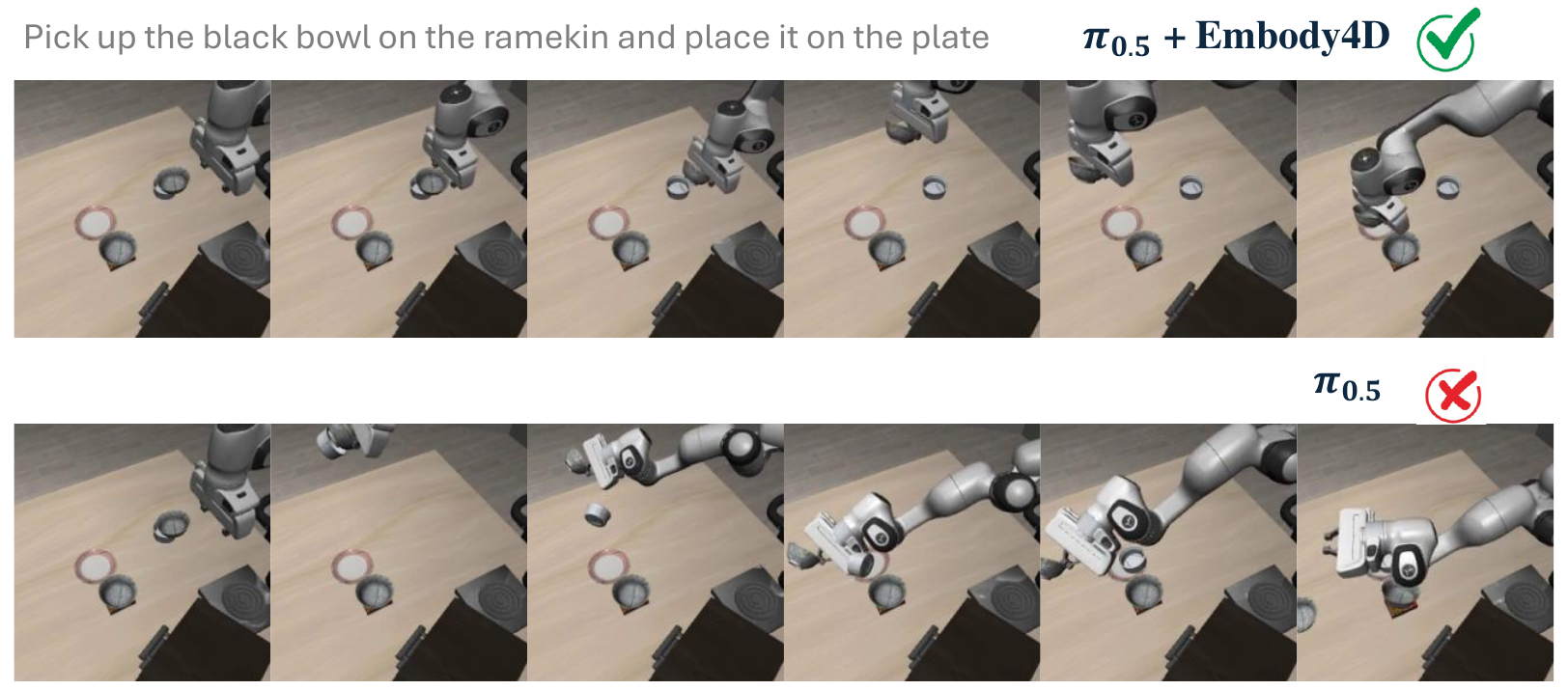}
    \caption{\textbf{An example episode from the LIBERO-Plus benchmark.} The policy fails in the original setting but successfully completes the task after the Embody4D is added.}
    \label{fig:compare}
    \vspace{-0.3cm}
\end{figure}

\subsection{Real World Experiments}

\begin{figure}[t]
    \raggedright
    \includegraphics[width=1\linewidth]{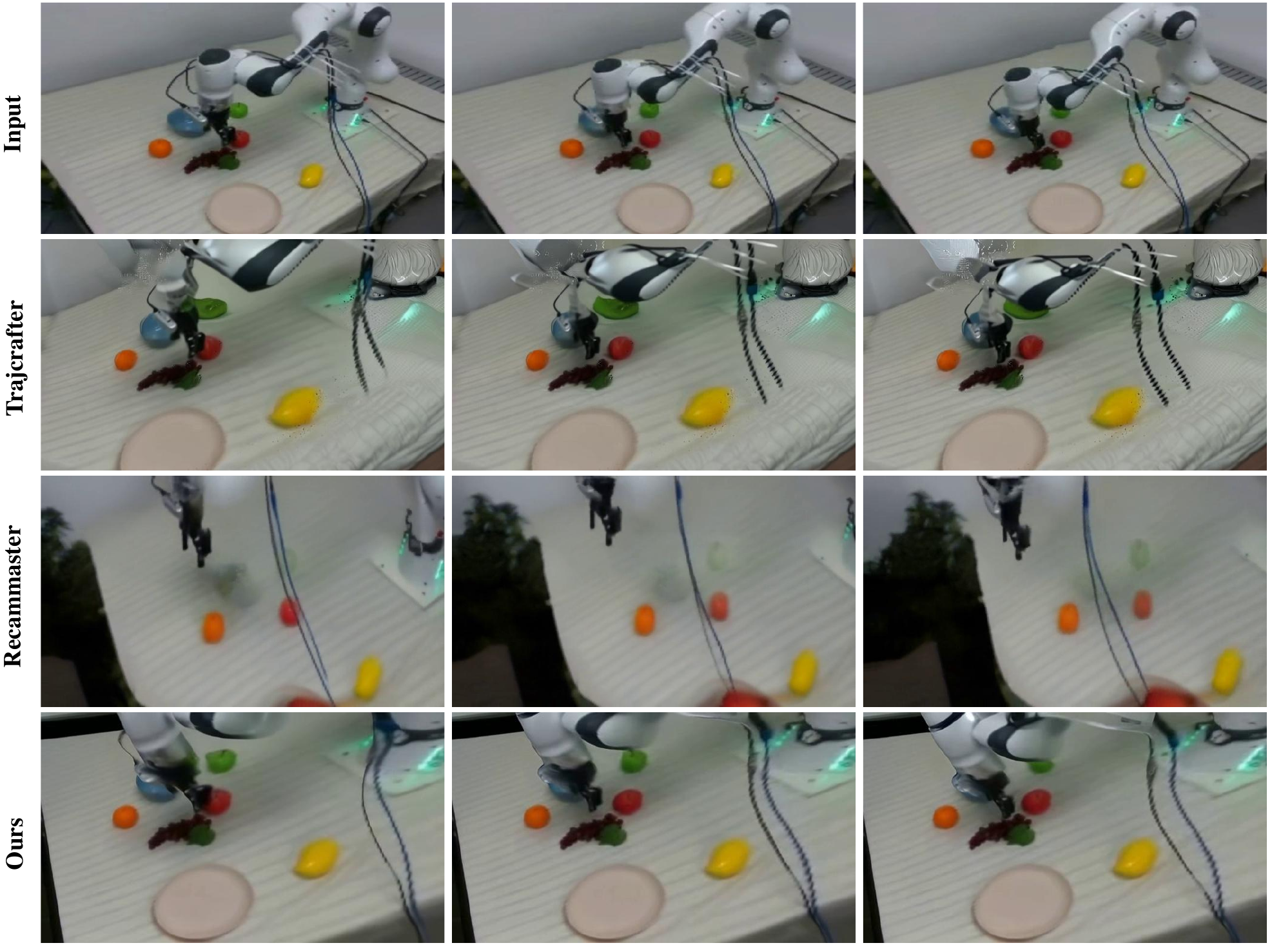}
    \caption{\textbf{Visualization results of real-world embodied experiments.}}
    \label{fig:demo}
\end{figure}

In our real-world experiments, the input camera pose is illustrated in Fig. \ref{fig:demo}. The target output pose is defined by a \ang{20} rightward rotation and a \SI{30}{\cm} forward translation relative to the input. We selected TrajectoryCrafter \cite{trajectorycrafter} and ReCamMaster \cite{recammaster}, the two top-performing models as baselines to generate videos from identical target viewpoints for a fair comparison. Our robotic manipulation setup employs a third-person view coupled with a wrist-mounted camera as the primary baseline. To evaluate performance improvements, we integrated novel-view data generated by Embody4D, TrajCrafter, and ReCamMaster into the training pipeline and measured their respective performance gains relative to the baseline. Consistent with the main text, our downstream model is the current state-of-the-art Vision-Language-Action model $\pi_{0.5}$. In addition, we collect one hundred human demonstration trajectories for post-training.

The experimental results are summarized in Tab. \ref{tab:realworld}. ReCamMaster, designed for continuous camera motion via direct extrinsic control, struggles with motion instability when simulating the fixed camera poses required by our downstream tasks, leading to a performance decline. TrajCrafter also delivers sub-optimal performance in embodied scenarios, often suffering from visual artifacts and hallucinations, resulting in marginal improvements. As shown in Fig. \ref{fig:demo}, incorporating a 4D generation model improves downstream manipulation and planning performance, with particularly significant gains on OOD tasks. Embody4D achieves better generation quality than the other two baseline models. In contrast, ReCamMaster relies on camera extrinsic control, which can easily introduce camera jitter, thereby leading to a decline in performance. Additionally, our method supports scalability by integrating multiple third-person image sequences into the VGGT \cite{vggt} pipeline, enabling a more comprehensive point cloud reconstruction that can be projected onto flexible target views for completion and novel view synthesis. 

As shown in Fig.~\ref{fig:vla}, Embody4D generates flexible novel-view observations from a single input view, serving both as policy-training augmentation and as auxiliary visual input during inference. While Embody4D is primarily formulated as video-to-video novel-view generation, it also supports image-to-image generation at inference time. Given a single-view image observation and a target relative viewpoint change, Embody4D synthesizes the corresponding novel-view image without requiring additional camera hardware.

\begin{figure}[h]
    \raggedright
    \includegraphics[width=1\linewidth]{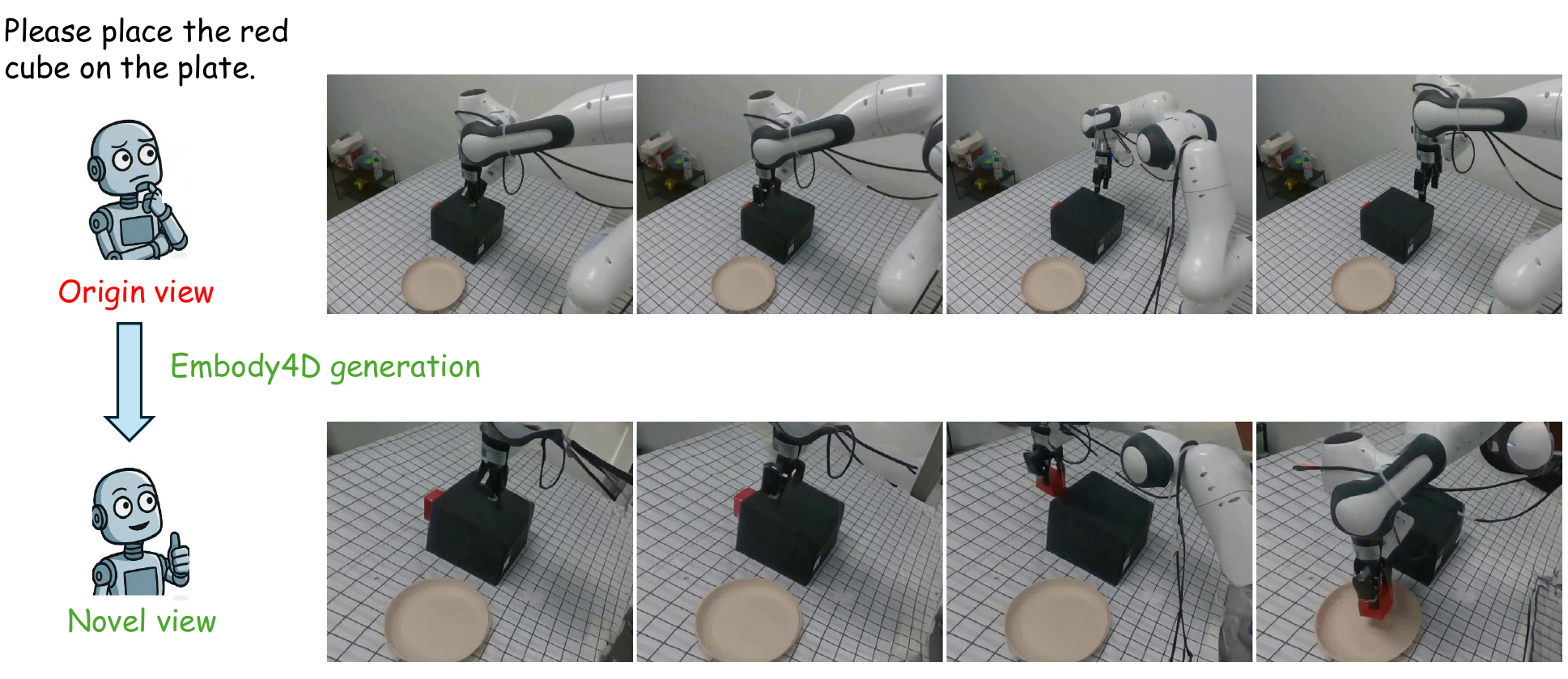}
    \caption{\textbf{Application of Embody4D during VLA inference.} In addition to video-to-video generation, Embody4D supports image-to-image novel-view generation during VLA inference, providing complementary visual observations that can reveal occluded objects and assist the VLA in generating more reliable actions.}
    \label{fig:demof}
\end{figure}

As illustrated in Fig.~\ref{fig:demof}, the upper part shows a failure case where the red cube is occluded in the original observation. Without a viewpoint change, the VLA receives insufficient visual evidence to localize the target object and therefore fails to complete the task. In contrast, the lower part shows that Embody4D synthesizes a complementary novel view without requiring additional camera hardware, exposing the previously occluded region and providing geometric and spatial cues for downstream reasoning. This additional observation enables the VLA to identify the red cube more reliably and generate the correct action.

\begin{figure}[t]
    \centering
    \begin{minipage}{1\linewidth}
        \centering
        \includegraphics[width=\linewidth]{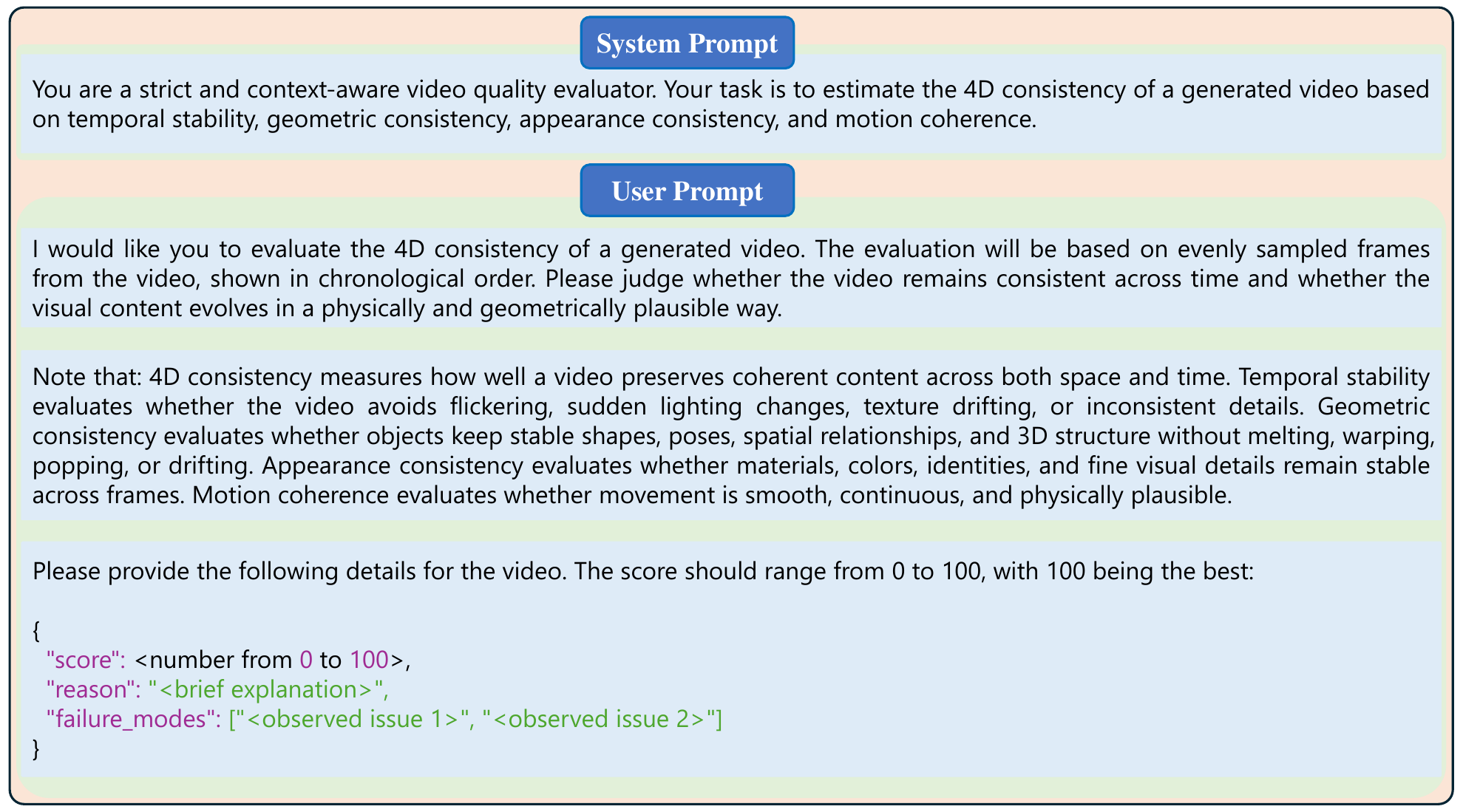}
    \end{minipage}
    \caption{\textbf{Prompt template for evaluating 4D video consistency.} }
    \label{fig:gpt}
    \vspace{-0.1cm} 
\end{figure}

\section{More Experimental Details}

\subsection{Dataset Curation}
\begin{figure}[!h]
    \centering
    \includegraphics[width=1\linewidth]{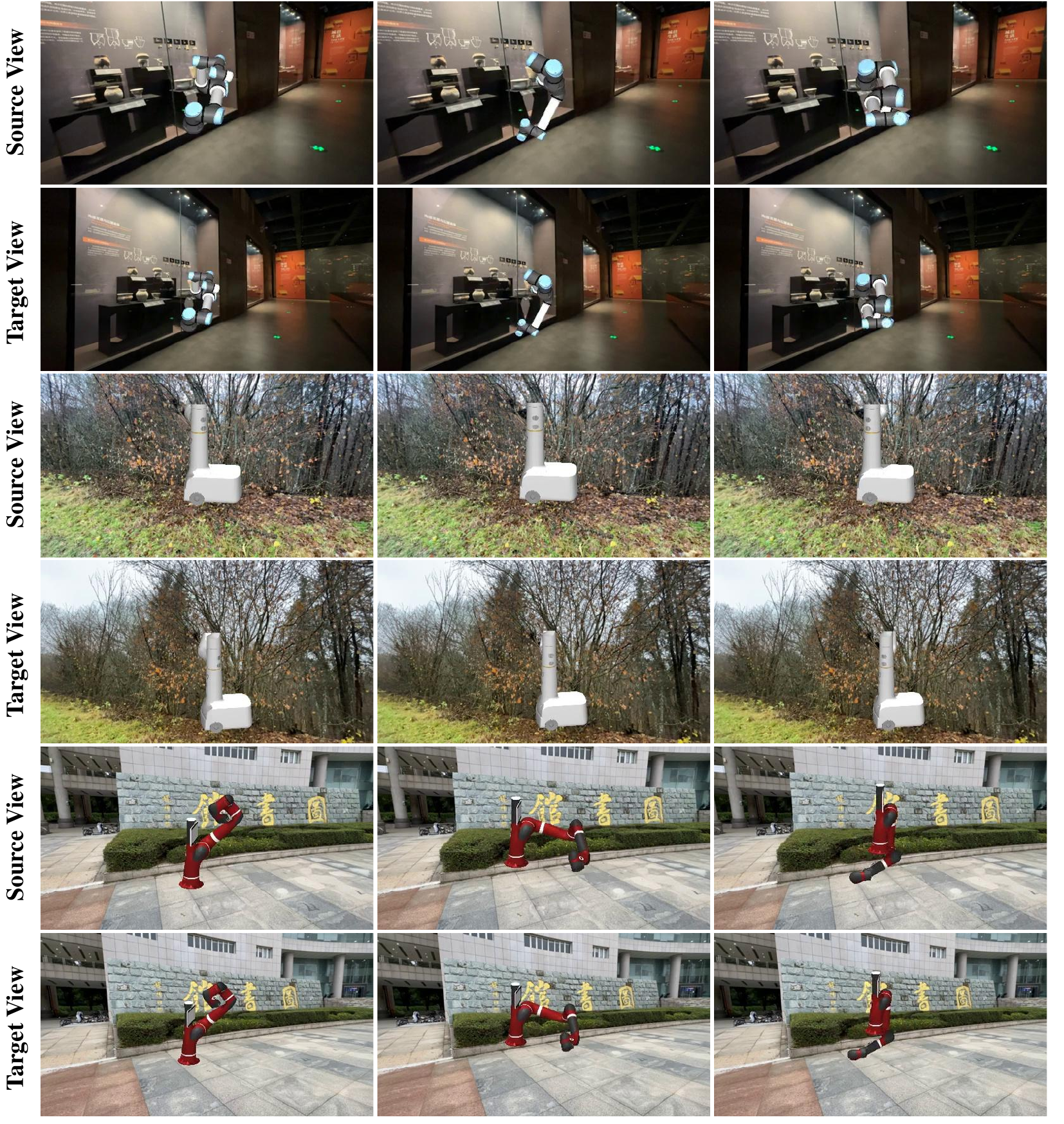}
    \caption{\textbf{Visualization results of compositional 4D embodied data}, featuring 4D-consistent sequences with diverse robotic arm foregrounds and varied backgrounds.}
    \vspace{-0.5cm}
    \label{fig:data}
\end{figure}
When rendering the robotic arm foreground in the MuJoCo \cite{mujoco} simulation environment, we ensure that the simulated motion closely mimics real world dynamics by strictly constraining the movement magnitude. Specifically, each movement step size is set to one hundredth of the total range of motion for each joint. To maintain spatial consistency, the camera placement within the simulation environment must strictly adhere to the extrinsic parameters of the target backgrounds intended for compositing. This process necessitates a precise coordinate transformation to map the OpenGL coordinate system used in the DL3DV  \cite{dl3dv} dataset into the XML format required by MuJoCo, which is defined by specific values for distance as well as azimuth and elevation angles.

Our empirical evaluation demonstrated that employing 49 frame sequences with independent camera and background poses leads to suboptimal performance. The compounding motion of both the foreground robotic arm and the background camera generates highly complex data that hinders the ability of the model to achieve convergence. To mitigate this issue, each data sample consists of two separate videos captured from two distinct and fixed camera poses. To avoid excessive parallax that may cause the composited robotic arm to drift out of the frame, we use GPT-4o~\cite{gpt} to select background sequences with moderate viewpoint variation. Specifically, for each scene in the DL3DV dataset, we randomly choose a starting frame and take the subsequent 40 frames as a temporally coherent background sequence. The selected sequence is required to preserve the same scene identity and dominant content, while exhibiting sufficient but not excessive parallax induced by camera motion. To this end, we design detailed prompts to filter sequences according to both geometric variation and semantic consistency. After foreground-background composition, we further apply an additional GPT-based filtering step to discard implausible or inconsistent composites, as illustrated in Fig.~\ref{fig:gpt}.

As illustrated in Fig. \ref{fig:data}, our training pairs are annotated by concatenating the identifier of the robotic arm with a detailed background description to form the final text prompt. This composition based approach yields more precise labels than the direct annotation of the final composited video sequence. For this purpose, we employ the CogVLM2 Llama3 Caption model \cite{cogvideo} as our primary labeling tool to ensure high quality descriptive metadata.

\subsection{Implementation Details of Latent Confidence-Aware Expert Modulation}

\begin{figure}[t]
    \raggedright
    \includegraphics[width=1\linewidth]{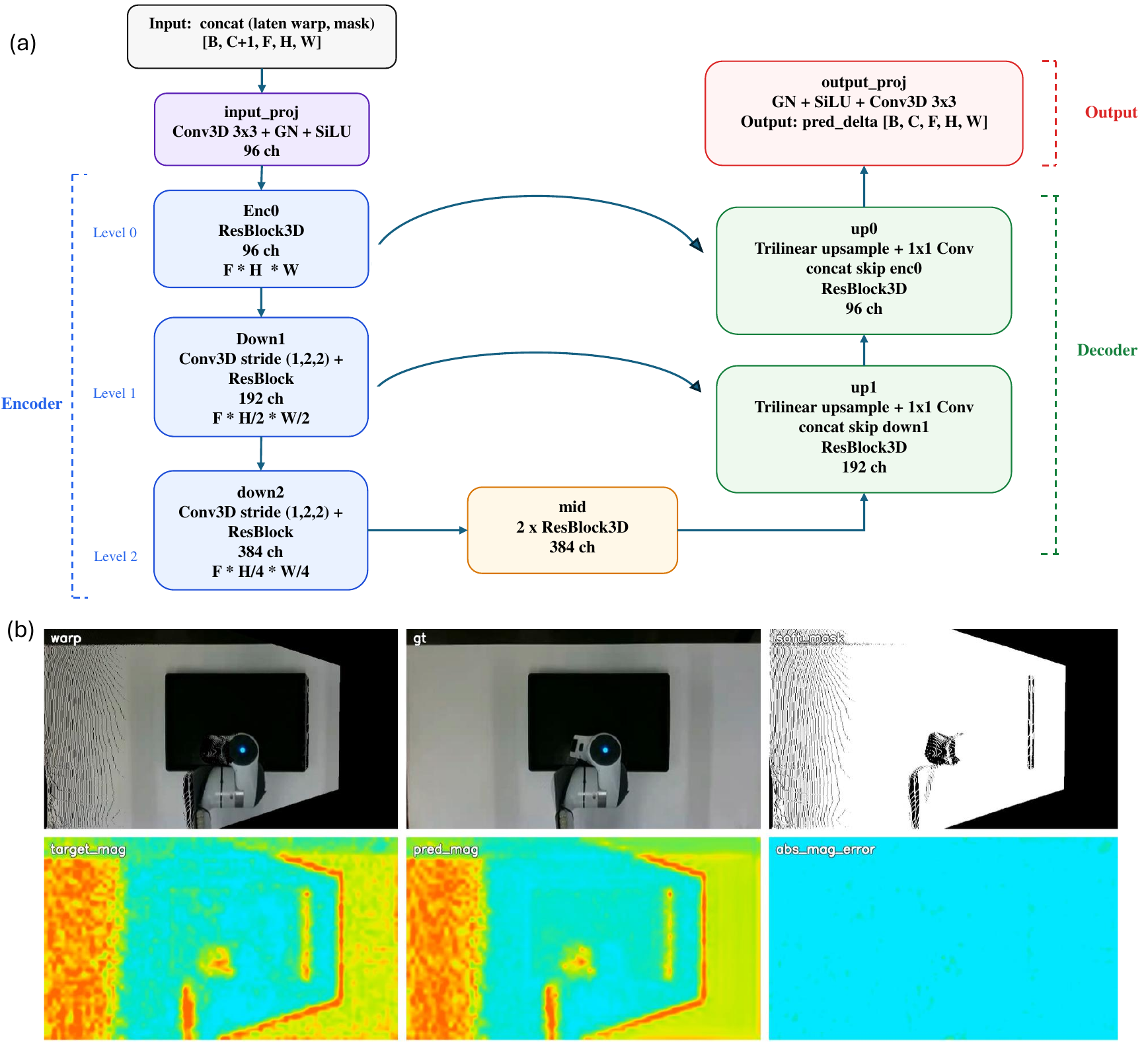}
    \caption{\textbf{Latent Confidence Estimation Module.} 
    (a) U-Net architecture for estimating latent confidence maps from the warped latent and soft validity mask. 
    (b) Visualization of the estimated confidence maps, where lower-confidence regions indicate higher correction demand for expert modulation.}
    \vspace{-0.1cm}
    \label{fig:latent}
\end{figure}
\paragraph{Latent Residual Routing.}
We design a lightweight 3D U-Net to estimate the latent discrepancy between the warped prior and the ground-truth target, which provides the routing signal for expert modulation. As shown in Fig.~\ref{fig:latent}, the warped latent is concatenated with the soft validity mask, resized to the latent resolution, and fed into the residual estimator. The estimator predicts the latent-space correction demand required to transform the warped prior toward the target. A larger predicted discrepancy indicates that the warped prior deviates more from the target latent and therefore requires stronger correction.

The residual estimator is trained on the full training set for two epochs and then frozen during expert training. As shown in Fig.~\ref{fig:latent}(b), the rightmost error map indicates that the estimator closely fits the target discrepancy. More importantly, the predicted discrepancy map exhibits a clear region-level structure, corresponding to different correction demands in the warped latent. This motivates us to route different latent regions to specialized FFN experts.

In 4D video generation, the target-view video is not generated freely from noise, but is jointly constrained by the source-view observation and geometric warping. Accordingly, regions in the warped latent have different semantic roles. Some regions remain highly consistent with the source-view observation and should be preserved; some regions are still supported by the source view but require fine-grained correction due to geometric errors, occlusion boundaries, or local misalignment; and other regions are newly exposed in the target view, where the source view cannot provide sufficient content and the model must synthesize plausible details from context. Therefore, the latent space can be naturally decomposed into three semantic regions: \emph{copy}, \emph{repair}, and \emph{inpaint}.

Based on this observation, we employ three FFN experts to model these three region types. This design matches the intrinsic structure of the task while avoiding unnecessary expert fragmentation. With fewer experts, the same parameters must handle both reliable source-supported regions and newly exposed unseen regions, creating a conflict between preserving source-view consistency and generating novel content. Conversely, using more experts may increase capacity but can make expert boundaries unstable and spatially fragmented, assigning adjacent regions to different experts and harming spatial continuity and temporal consistency. The three-expert design therefore provides a balanced trade-off between semantic specialization and routing stability.

We then define a confidence-aware latent routing function. Let \(\bar{V}\) denote the smoothed source coverage map derived from the soft validity mask, and let \(\bar{D}\) denote the smoothed predicted residual magnitude. The validity mask mainly reflects whether a region is covered by source-view information, while the predicted residual magnitude further measures how much the warped latent deviates from the target latent. Therefore, regions with high source coverage but large residuals are routed to \emph{repair} rather than directly copied, whereas regions with low source coverage and large residuals are more likely to require \emph{inpainting}.

Specifically, copy regions are required to have both high source-view coverage and low residual magnitude. Inpaint regions correspond to positions with insufficient source-view coverage and high residual magnitude, or to severely misaligned positions with extremely large residuals. All remaining positions are assigned to the repair expert. We use the following routing thresholds:
\[
\tau_v=0.65,\quad
\tau_c=0.50,\quad
\tau_i=0.60,\quad
\tau_{iv}=0.50,\quad
\tau_f=0.80.
\]
Here, \(\tau_v\) denotes the minimum source coverage required for copy, \(\tau_c\) denotes the residual upper bound for copy, \(\tau_i\) denotes the residual threshold for assigning low-coverage regions to inpaint, \(\tau_{iv}\) denotes the low-coverage threshold, and \(\tau_f\) denotes the residual threshold for forced inpainting. The final routing rule is
\[
\mathcal{R}_{\mathrm{copy}}
=
\{\bar{V}\ge0.65 \land \bar{D}<0.50\},
\]
\[
\mathcal{R}_{\mathrm{inpaint}}
=
\{\bar{V}<0.50 \land \bar{D}\ge0.60\}
\cup
\{\bar{D}\ge0.80\},
\]
\[
\mathcal{R}_{\mathrm{repair}}
=
\Omega\setminus
(\mathcal{R}_{\mathrm{copy}}\cup\mathcal{R}_{\mathrm{inpaint}}).
\]
The thresholds are determined by a two-stage calibration procedure. First, we visualize route maps under different threshold combinations on randomly sampled videos to identify a stable candidate range. This step rejects degenerate settings: overly conservative copy thresholds shrink copy regions and weaken source-view constraints, while overly aggressive inpaint thresholds expand inpaint regions or introduce high-frequency routing fragments, causing the model to prematurely discard warped information. Second, we perform route profiling over the full training set and analyze the copy/repair/inpaint distributions across different data sources and disparity conditions. The final thresholds are selected to satisfy three criteria: copy regions should remain non-trivial even under large viewpoint changes, high-coverage regions with large residuals should be primarily routed to repair to reduce generative drift, and low-coverage or severely misaligned regions should be consistently routed to inpaint to preserve the ability to synthesize newly exposed content.

To reduce fragmentation caused by high-frequency latent residual responses, we apply local smoothing to the residual magnitude and source coverage map with \(9\times9\) and \(7\times7\) kernels, respectively. We further adopt top-2 expert routing, where the auxiliary expert weight is set to \(\alpha=0.20\), followed by a \(5\times5\) local smoothing on the route weights. This setting is selected as a trade-off between visual continuity and global route profiling: it removes isolated routing artifacts while preserving coherent inpaint regions.

\subsection{Acquisition of Interaction Priors}
\begin{figure}[!h]
    \centering
    \includegraphics[width=1\linewidth]{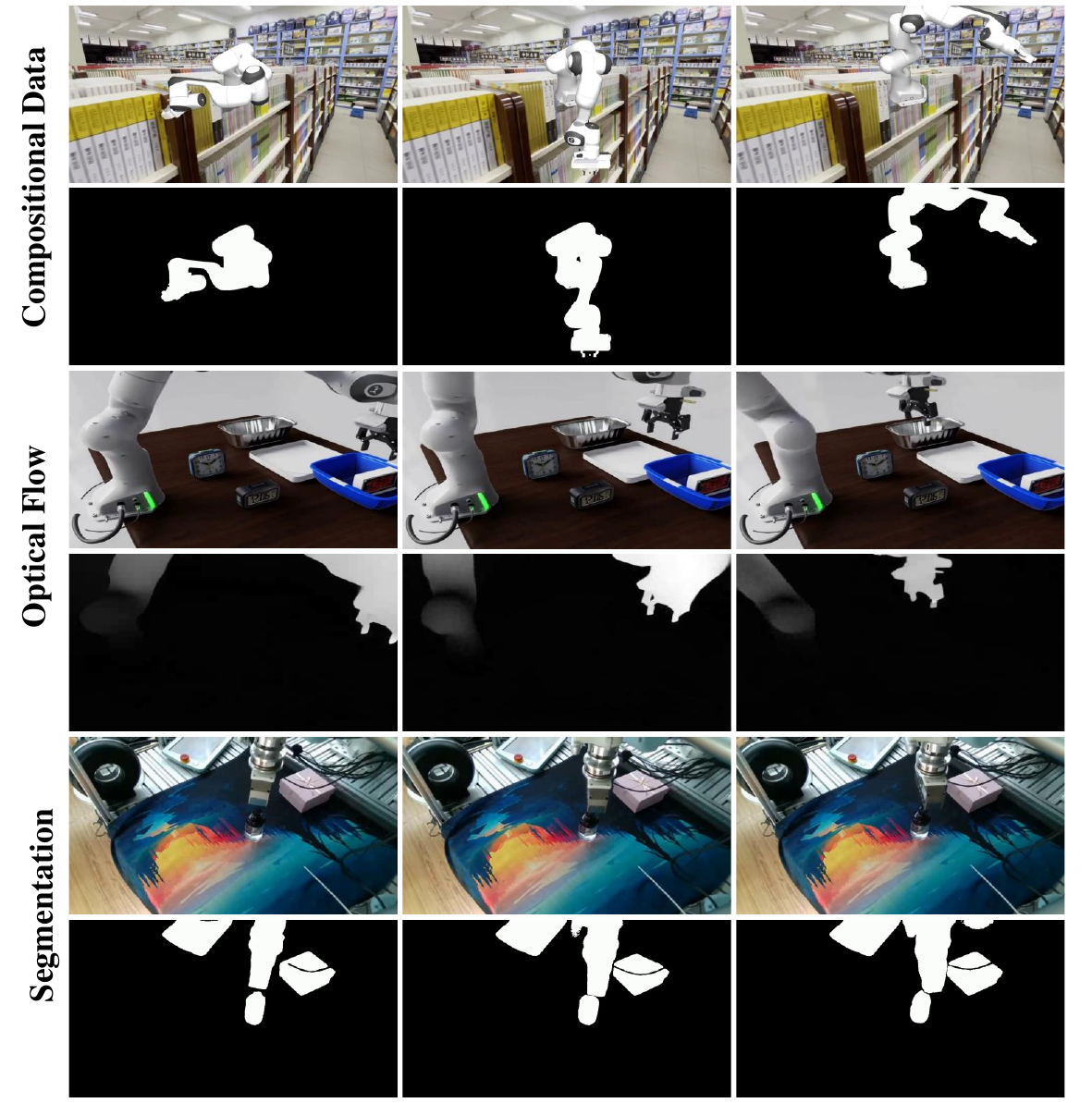}
    \caption{\textbf{Visualization of interaction motion priors.} Showcasing three different foreground acquisition strategies: direct acquisition from compositional data, optical flow estimation, and video object segmentation.}
    \vspace{-0.3cm}
    \label{fig:front}
\end{figure}

When utilizing the interaction-aware attention mechanism, it is first necessary to obtain a motion prior bias. For data captured in real world scenarios, camera movement often introduces a certain degree of jitter, which in turn leads to noise in the optical flow estimation \cite{memflow}. In cases where the noise level is excessively high, we directly employ the Segment Anything Model 3  \cite{sam3} to perform segmentation. The input prompt for this process is defined as ``robot arm and objects''. As illustrated in Fig. \ref{fig:front}, we provide several examples of the resulting optical flow estimations alongside the segmentation masks generated by the Segment Anything Model 3, all masks will be normalized to the range of 0 to 1 for calculations.
To prevent the model from over-relying on perfectly foregrounds during training, we introduce a dropout mechanism on the interaction masks. This forces the network to learn robust attention patterns even when the motion prior or mask quality is imperfect, ensuring that the model can handle real-world scenarios with noisy or partially missing foreground information.

\subsection{Training Implementation Details}
We implement our Embody4D using the TrajCrafter diffusion transformer as the generative backbone. All parameters are fully fine-tuned in an end-to-end manner. During training, the frame resolution is fixed at $384 \times 672$, with a sequence length of 49 frames.

We adopt a two-stage training strategy. In the first stage, the model is trained on our self-constructed combinatorial real-world 4D dataset to enhance temporal consistency. In the second stage, we freeze the interaction-aware blocks and utilize pseudo-4D data, generated by projecting interaction data onto flexible perspectives and back-projecting them to the initial view. This augmentation enables the model to generalize effectively to complex real-world interactions. Finally, all model parameters are fully fine-tuned in an end-to-end manner.

The training configurations are summarized as follows:
\begin{itemize}
    \item \textbf{Stage 1:} Conducted for 5.3k iterations with a learning rate of $1 \times 10^{-5}$.
    \item \textbf{Stage 2:} Conducted for 8.1k iterations with a learning rate of $2 \times 10^{-6}$.
\end{itemize}
Both stages utilize a mini-batch size of 16 (2 per GPU) and are performed on 8 NVIDIA A100 GPUs.
Moreover, the latent confidence-aware expert modulation adopts a soft routing
scheme and remains frozen during training. Based on the estimated confidence, each token is assigned
to two neighboring experts, and their responses are linearly combined with
weights $0.8$ and $0.2$ to form the final modulation signal. This avoids abrupt
expert switching and encourages smoother transitions across copy, repair, and
inpaint behaviors.

\section{Limitations and Future Works}
\begin{figure}[t]
    \hspace{-0.4cm}
    \raggedright
    \includegraphics[width=1\linewidth]{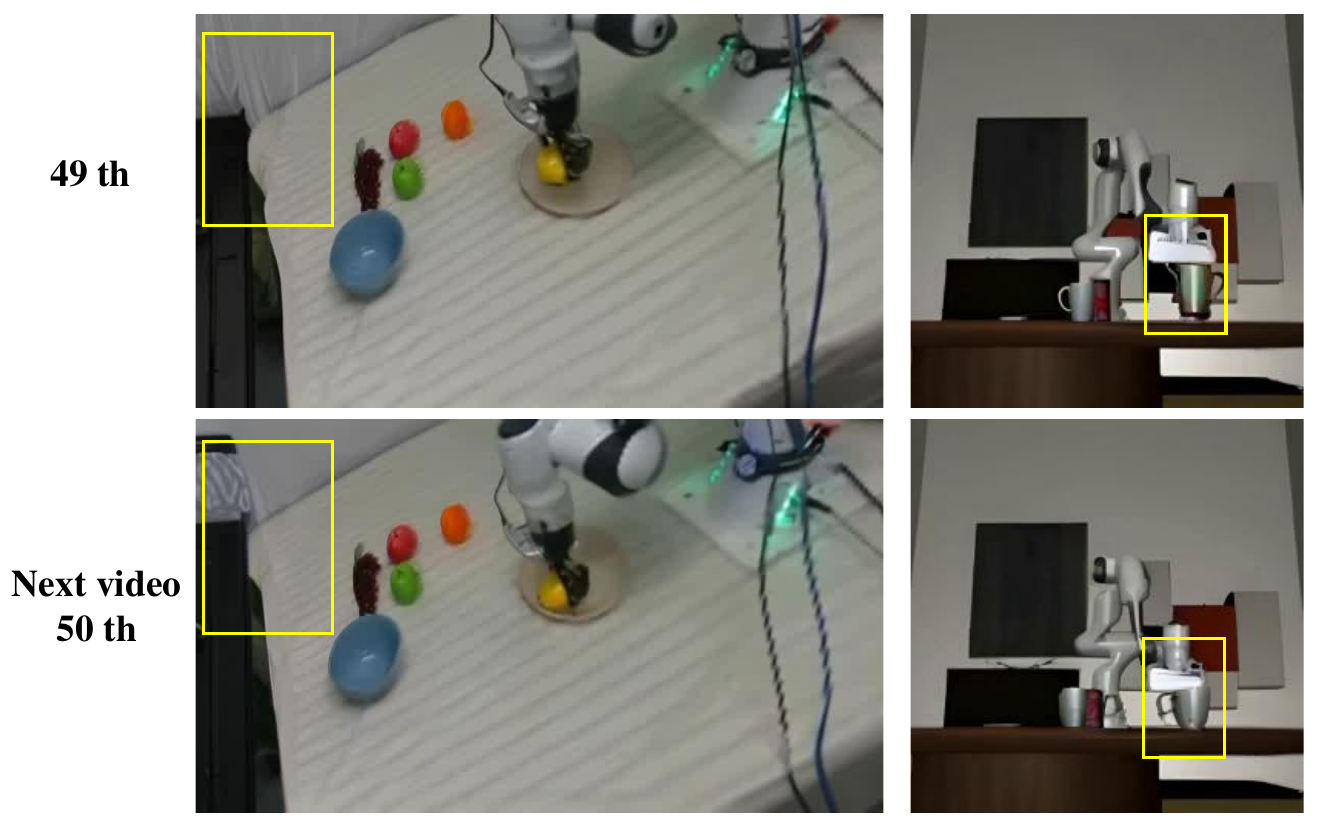}
    \caption{\textbf{Visualization of failure cases.}}
    \label{fig:failure}
\end{figure}

Due to the limited temporal modeling capability of current video generation foundation models, our framework can only synthesize fixed-length video sequences. For long input videos, we therefore split them into multiple short clips, perform novel-view synthesis independently for each clip, and then concatenate the generated clips into a complete sequence. However, since different clips are generated independently, temporal discontinuities or visual state inconsistencies may arise at clip boundaries, leading to mild jitter in the final video. Representative failure cases are shown in Fig. \ref{fig:failure}. Existing long-video generation techniques remain imperfect and still cannot reliably support arbitrary-length or indefinitely long video synthesis. In future work, we will explore more advanced video generation foundation models to improve long-horizon temporal modeling and cross-clip consistency.

\section{Supplementary Results}
\begin{figure}[h]
    \hspace{-0.4cm}
    \raggedright
    \includegraphics[width=1\linewidth]{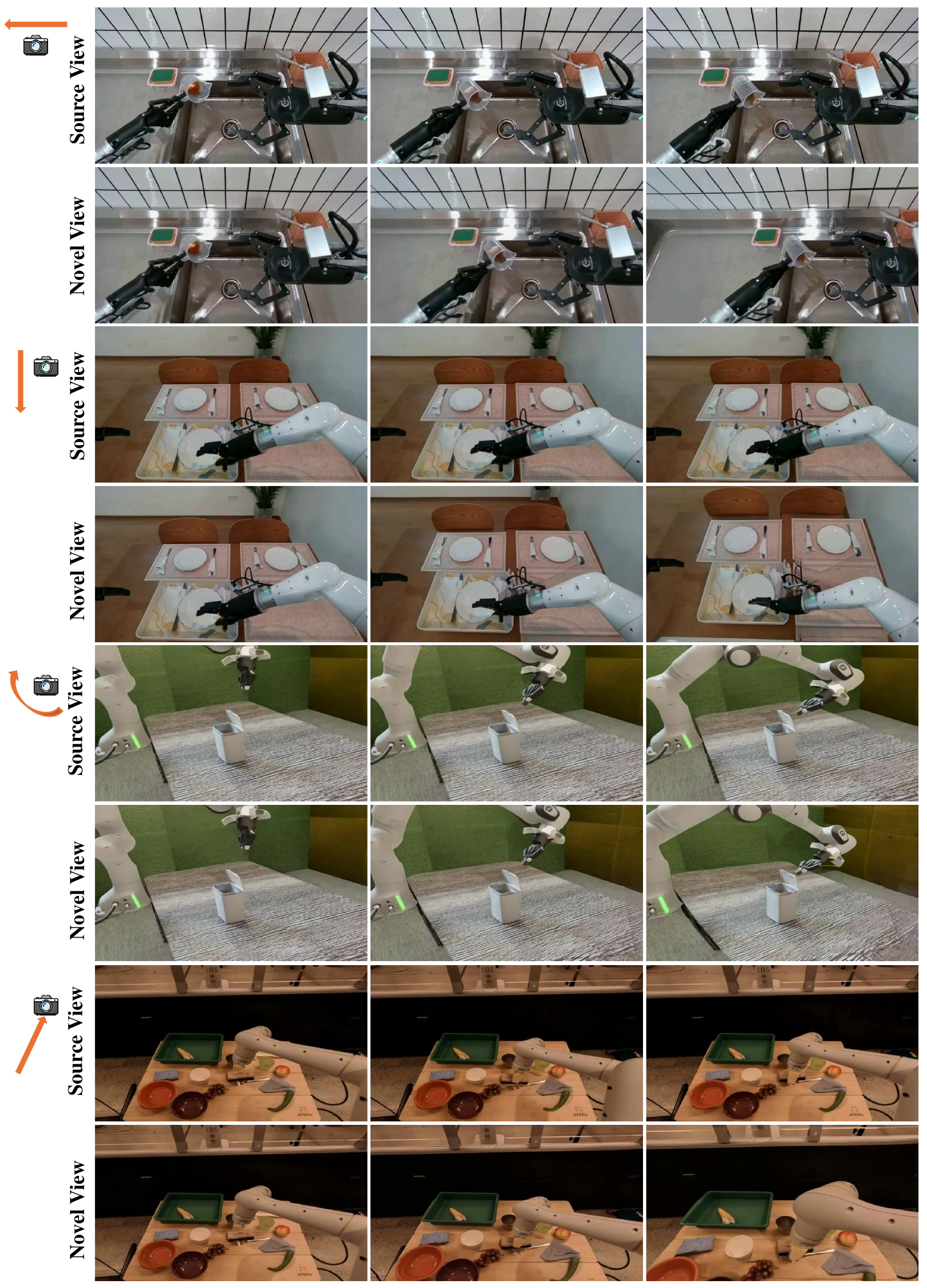}
    \caption{\textbf{Visualization results of Embody4D}}
    \label{fig:result1}
\end{figure}

\begin{figure}[h]
    \hspace{-0.4cm}
    \centering
    \includegraphics[width=1\linewidth]{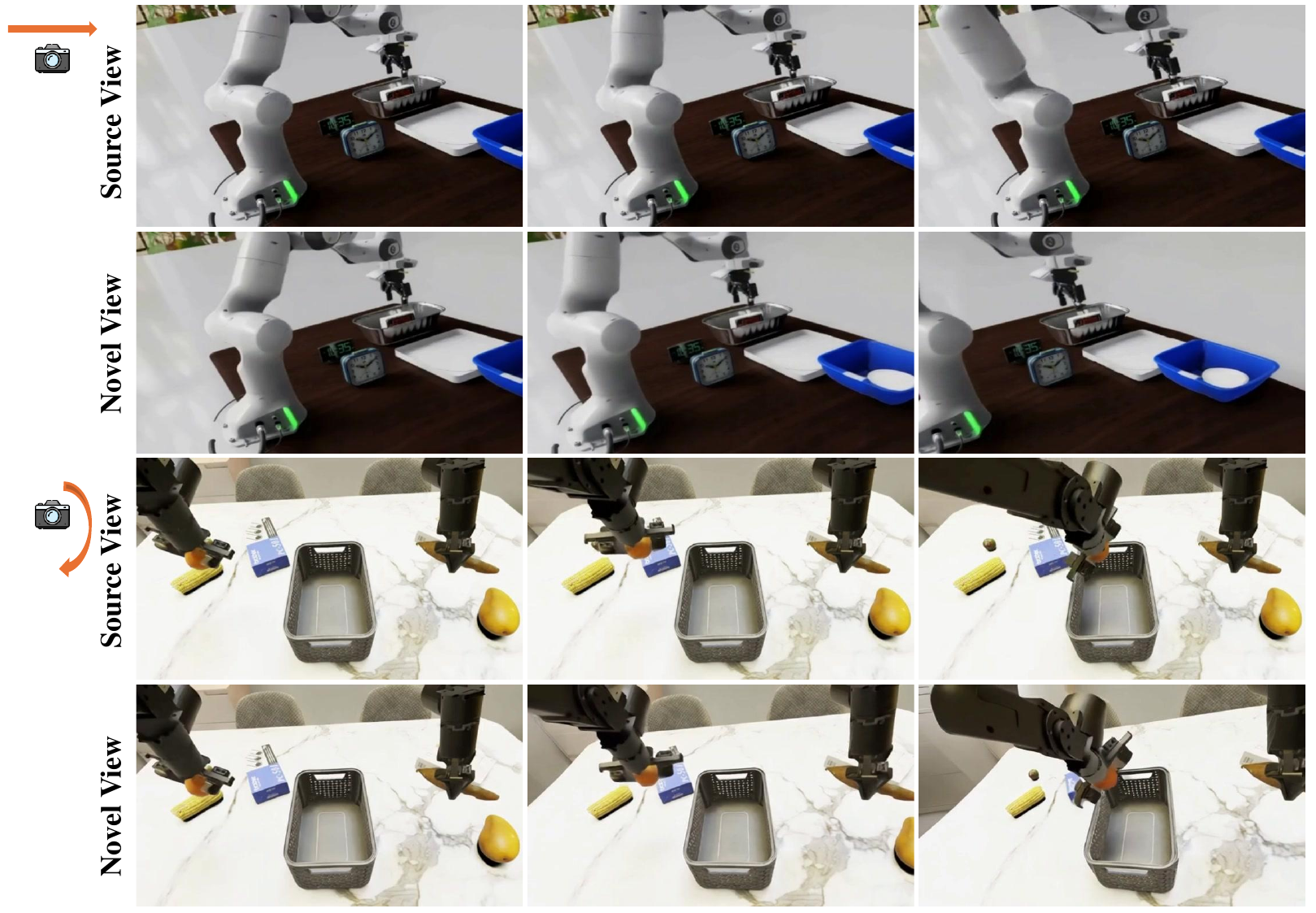}
    \caption{\textbf{Visualization results of Embody4D}}
    \vspace{-0.3cm}
    \label{fig:result2}
\end{figure}

\begin{figure}[t]
    \hspace{-0.4cm}
    \raggedright
    \includegraphics[width=1\linewidth]{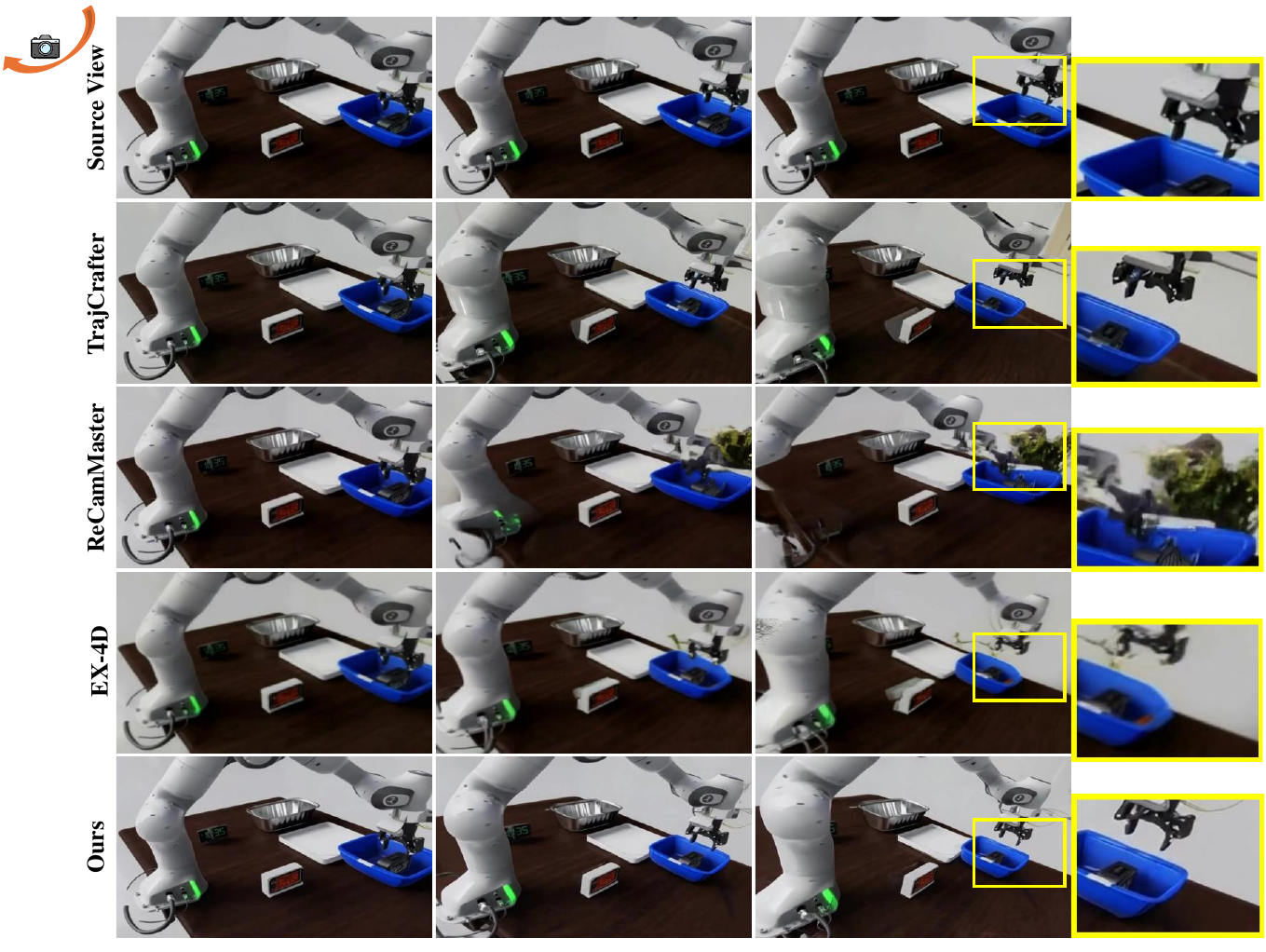}
    \caption{\textbf{Qualitative visualization results of our method and the baselines.}}
    \vspace{-0.3cm}
    \label{fig:result3}
\end{figure}

\begin{figure}[h]
    \hspace{-0.4cm}
    \raggedright
    \includegraphics[width=1\linewidth]{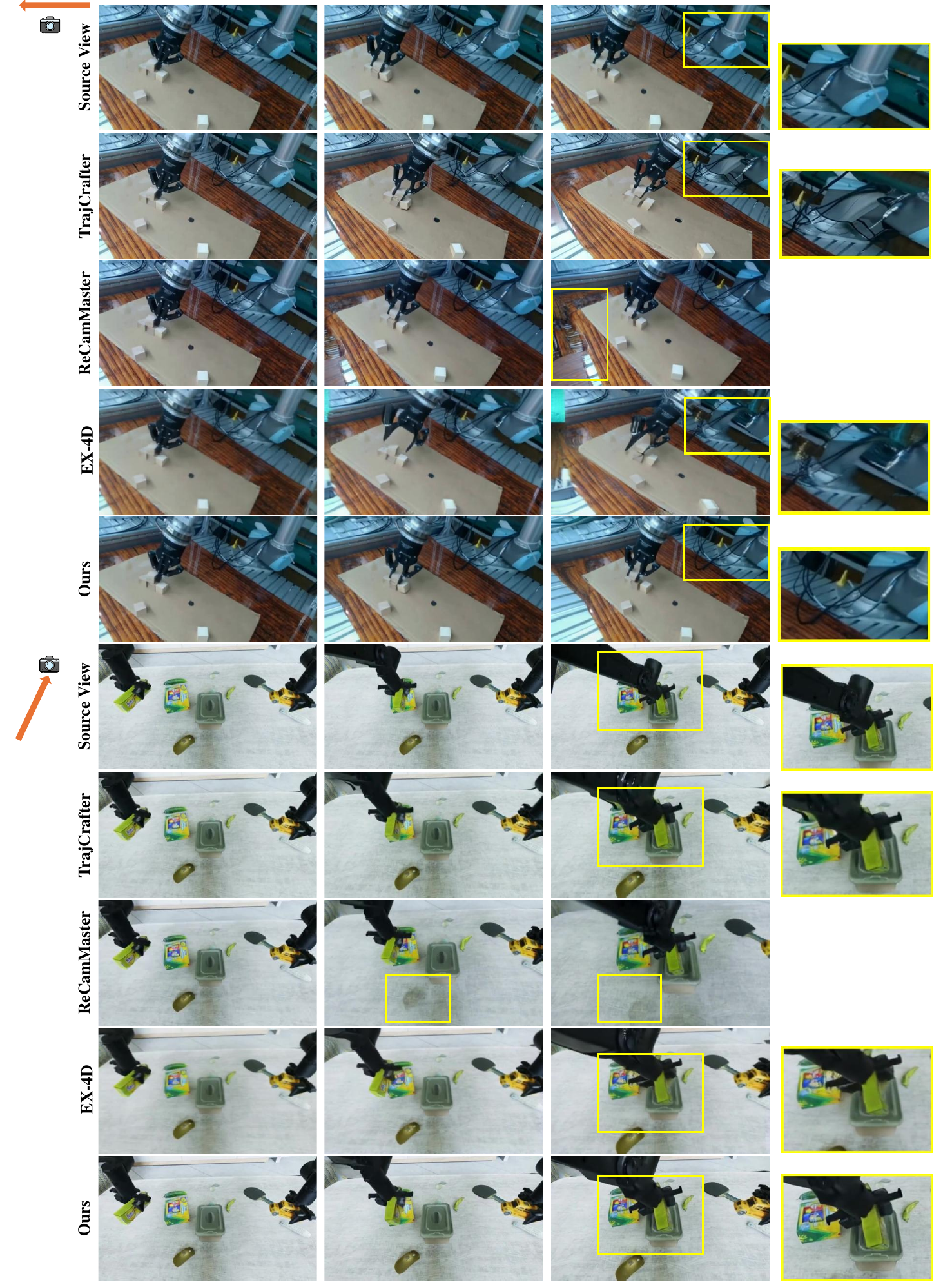}
    \caption{\textbf{Qualitative visualization results of our method and the baselines.}}
    \label{fig:result4}
\end{figure}

As shown in Fig. \ref{fig:result1}, Fig. \ref{fig:result2}, Fig. \ref{fig:result3}, and Fig. \ref{fig:result4}, we provide more experimental results and additional visual comparisons with the baseline method. These results highlight that our method achieves state-of-the-art quality, 
characterized by fine-grained interaction details, high visual fidelity, and robust spatiotemporal consistency.

\end{document}